\documentclass{article}

\PassOptionsToPackage{numbers, compress}{natbib}

    \usepackage[preprint]{neurips_2021}

\usepackage[utf8]{inputenc} 
\usepackage[T1]{fontenc}    
\usepackage{hyperref}       
\usepackage{url}            
\usepackage{booktabs}       
\usepackage{amsfonts}       
\usepackage{nicefrac}       
\usepackage{microtype}      
\usepackage{xcolor}         
\usepackage{graphicx}
\usepackage{booktabs}
\usepackage{subfig}
\usepackage{wrapfig}

\usepackage[linesnumbered,ruled]{algorithm2e}
\IncMargin{1em}

\SetCommentSty{mycommfont}

\title{Cross-Model Consensus of Explanations and Beyond for Image Classification Models: An Empirical Study}

\author{%
    Xuhong Li \\
    Baidu Research, Baidu Inc \\
    \texttt{lixuhong@baidu.com}
    \And
    Haoyi Xiong \\
    Baidu Research, Baidu Inc \\
    \texttt{xionghaoyi@baidu.com}
    \And
    Siyu Huang \\
    Nanyang Technological University \\
    \texttt{siyu.huang@ntu.edu.sg}
    \And
    Shilei Ji \\
    Baidu Inc \\
    \texttt{jishilei@baidu.com}
    \And
    Dejing Dou \\
    Baidu Research, Baidu Inc \\
    \texttt{doudejing@baidu.com}
}

\usepackage{amsmath,amsfonts,bm}

\def\eqref#1{equation~\ref{#1}}

\def\1{\bm{1}}

\def\va{{\bm{a}}}
\def\vb{{\bm{b}}}
\def\vc{{\bm{c}}}

\def\mL{{\bm{L}}}
\def\mM{{\bm{M}}}

\DeclareMathAlphabet{\mathsfit}{\encodingdefault}{\sfdefault}{m}{sl}
\SetMathAlphabet{\mathsfit}{bold}{\encodingdefault}{\sfdefault}{bx}{n}

\def\gA{{\mathcal{A}}}

\def\gD{{\mathcal{D}}}

\def\gM{{\mathcal{M}}}

\begin{document}

\maketitle

\begin{abstract}
    
    %
    
    Existing interpretation algorithms have found that, even deep models make the same and right predictions on the same image, they might rely on different sets of input features for classification. However, among these sets of features, some common features might be used by the majority of models. In this paper, \textit{we are wondering what are the common features used by various models for classification and whether the models with better performance may favor those common features}. For this purpose, our works uses an interpretation algorithm to attribute the importance of features (e.g., pixels or superpixels) as explanations, and proposes the \emph{cross-model consensus of explanations} to capture the common features. Specifically, we first prepare a set of deep models as a \emph{committee}, then deduce the explanation for every model, and obtain the \textit{consensus} of explanations across the entire committee through \textit{voting}.
    With the cross-model consensus of explanations, we conduct extensive experiments using 80+ models on 5 datasets/tasks. We find three interesting phenomena as follows: (1) the consensus obtained from image classification models is aligned with the ground truth of semantic segmentation; (2) we measure the similarity of the explanation result of each model in the committee to the consensus (namely \textit{consensus score}), and find positive correlations between the consensus score and model performance; and (3) the consensus score coincidentally correlates to the interpretability. 

\end{abstract}

\section{Introduction}

    Deep models are well-known by their excellent performance in many challenging domains, as well as their black-box nature. To interpret the prediction of a deep model, a number of trustworthy interpretation algorithms~\citep{bach2015pixel,zhou2016learning,ribeiro2016should,smilkov2017smoothgrad,sundararajan2017axiomatic,lundberg2017unified} have been recently proposed to attribute the importance of every input feature in a given sample with respect to the model's output.
    %
    For example, given an image classification model, LIME~\cite{ribeiro2016should} and SmoothGrad~\cite{smilkov2017smoothgrad} could attribute the importance scores to every superpixel/pixel in an image with respect to the model's prediction. In this way, one can easily explain the classification result of a model with a data point by visualizing the important features used by the model for prediction.
    
    The use of interpretation tools finds that, even deep models make the same and right predictions on the same image, they might rely on different sets of input features for classification. For example, our work uses LIME and SmoothGrad to explain a number of models trained on image classification tasks on the same set of images and obtains different explanations for these models even all they make right predictions (latterly shown in Figure~\ref{fig:imagenet_models_qualitative} and Figure~\ref{fig:cub_models_qualitative}). While these models have been explained to make the same prediction using different sets of features, we can still find that some common features might be used by the majority of models. 
    In this way, we are particularly interested in two research questions as follows: {(1)} \emph{What are the common features used by various models in an image?} {(2)} \emph{Whether the models with better performance favor those common features?}

\begin{figure*}[t]
		\centering
		{\includegraphics[width=0.95\linewidth]{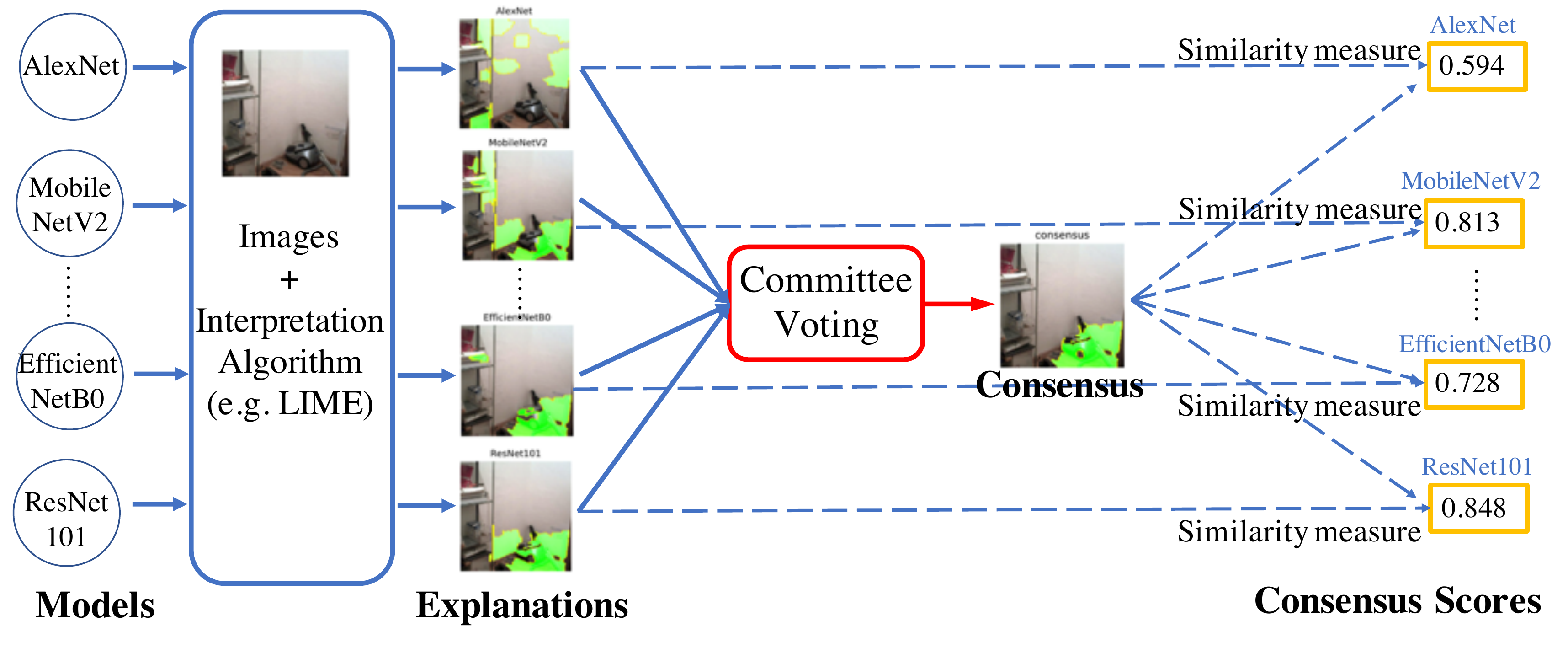}}
        \caption{Illustration of the proposed framework that consists of the three steps: (1) prepares a set of trained models as committee, (2) aggregates explanation results across the committee to get the consensus, and (3) computes the similarity score of each explanation to the consensus.}
		\label{fig:pipeline}
	\end{figure*}

    To answer these two questions, we propose to study the common features across a number of deep models and measure the similarity between the set of common features and the one used by every single model. Specifically, as illustrated in Figure~\ref{fig:pipeline}, we generalize an electoral system to first form a \textit{committee} with a number of deep models, then obtain the explanations for a given image based on one trustworthy interpretation algorithm, then call for \textit{voting} to obtain the \textit{cross-model consensus of explanations}, or shortly \textit{consensus}, and finally compute a similarity score between the consensus and the explanation result for each deep model, denoted as \textit{consensus score}. Through extensive experiments using 80+ models on 5 datasets/tasks, we find that (1) the consensus is aligned with the ground truth of image semantic segmentation; (2) a model in the committee with a higher consensus score usually performs better in terms of testing accuracy; and (3) models' consensus scores coincidentally correlates to their interpretability. 

    
    The contributions of this paper could be summarized as follows. To the best of our knowledge, this work is the first to investigate the common features, which are used and shared by a large number of deep models for image classification, by incorporating with interpretation algorithms. We propose the cross-model consensus of explanations to characterize the common features, and connect the consensus score to the performance and interpretability of a model. Finally, we obtain three observations from the experiments with thorough analyses and discussions.

	\section{Related Work}
	
	We first review the interpretation algorithms and the evaluation approaches on their trustworthiness.
	To visualize the activated subregions of intermediate-layer feature maps, many algorithms have been proposed to interpret convolutional networks~\cite{zhou2016learning,selvaraju2020grad,chattopadhay2018grad,wang2020score}.
	Apart from investigating the inside of complex deep networks, simple linear or tree-based surrogate models have been used as ``out-of-box explainers'' to explain the predictions made by the deep model over the dataset through local or global approximations~\cite{ribeiro2016should,van2019global,ahern2019normlime,zhang2019interpreting}.
	Instead of using surrogates for deep models, algorithms, such as SmoothGrad \citep{smilkov2017smoothgrad}, Integrated Gradients \citep{sundararajan2017axiomatic}, DeepLIFT \citep{shrikumar2017learning} etc, have been proposed to estimate the input feature importance with respect to the model predictions.
	Note that there are many other interpretation algorithms and we mainly discuss the ones that are related to feature attributions and suitable for deep models for image classification in this paper.
	Evaluations on the trustworthiness of interpretation algorithms are of objective to qualify their trustworthiness and not mislead the understanding of models' behaviors, e.g. \citet{adebayo2018sanity} have found that some algorithms are independent both of the model and the data generating process, through randomizing the parameters of models.
	Other evaluation approaches include perturbation of important features~\cite{samek2016evaluating,Petsiuk2018rise,vu2019evaluating,hooker2019benchmark}, model trojaning attacks~\cite{chen2017targeted,gu2017badnets,lin2020you}, infidelity and sensitivity~\cite{ancona2018towards,yeh2019fidelity} to similarity samples in the neighborhood, through a crafted dataset~\cite{yang2019benchmarking}, and user-study experiments~\cite{lage2019evaluation,jeyakumar2020can}.
	
	From an orthogonal perspective, evaluations across models are also urged for building more interpretable and explainable AI systems.
	However, evaluations across the deep models are scarce.
	\citet{bau2017network} proposed \textit{Network Dissection} to build an additional dataset with dense annotations of a number of visual concepts for evaluating the interpretability of convolutional neural networks.
    Given a convolutional model, Network Dissection recovers the intermediate-layer feature maps used by the model for the classification, and then measures the overlap between the activated subregions in the feature maps with the densely human-labeled visual concepts to estimate the interpretability of the model.
    Another common solution to the evaluation across deep models is user-study experiments~\citep{doshi2017towards}.
    
    In this paper, we do not directly evaluate the interpretability across deep models, but based on the proposed framework, we show experimentally that the consensus score is positively correlated to the generalization performance of deep models and coincidentally related to the interpretability.
    We will discuss more details with analyses later.
    We believe that based on the explanations, our proposed framework and the consensus score could help to better understand deep models.

	\section{Framework of Cross-Model Consensus of Explanations}
	\label{sec:consensus}
	


    \begin{wrapfigure}[21]{R}{0.6\linewidth}
	\begin{algorithm}[H]
	    \DontPrintSemicolon
        \caption{Framework Pseudocode.}
        
	    \label{algo:consensus-short}
		
		\textbf{Input:} A dataset $\gD$ and an interpretation algorithm $\gA$.
		
		\tcc{ Step 1: Committee Formation with Deep Models}
        Train $m$ deep models on $\gD$ that form the committee $\gM$.
        
        \tcc{ Step 2: Committee Voting for Consensus Achievement}
		For each example $d_i$ in $\gD$, initialize an empty matrix $\mL$ for storing the explanations.
		
		\For{\underline{each model} $\mM_j$ in $\gM$}
         {$\mL_j=\text{interpret}(\gA, d_i, \mM_j)$ 
        }
        
        $\vc = \text{reach\_consensus}(\mL)$ 
        
        \tcc{ Step 3: Consensus-based Similarity Score}
        
        \For{\underline{each model} $\mM_j$ in $\gM$}{
        $s_j=\text{similarity}(\mL_j, \vc)$ as the score of $\mM_j$ for $d_i$.
        }
        
        Repeat \textbf{Step2} and \textbf{Step3} for all examples in $\gD$. For each model $\mM_j$, the overall consensus score is the average of the similarity scores over all examples.
        
    \end{algorithm}
	\end{wrapfigure}
	
	In this section, we introduce the proposed approach that generalizes the electoral system to provide the consensus of explanations across various deep models. 
	Specifically, the proposed framework consists of three steps, as detailed in the following.
	
	\textbf{Step1: Committee Formation with Deep Models.} Given $m$ deep models that are trained for solving a target task (image classification task in our experiments) on a visual dataset where each image contains one main object, the approach first forms the given deep models as a \textit{committee}, noted as $\gM$, and then considers the variety of models in the committee that would establish the consensus for comparisons and evaluations.
	
	\textbf{Step2: Committee Voting for Consensus Achievement.}
	With the committee of deep models and the task for explanation, the proposed framework leverages a trustworthy interpretation tool $\gA$, e.g. we choose LIME~\citep{ribeiro2016should} or SmoothGrad~\citep{smilkov2017smoothgrad} as $\gA$ in this paper, to obtain the explanation of every model on every image in the dataset. Given some sample $d_i$ from the dataset, we note the obtained explanation results of all models as $\mL$. Then, we propose a \textit{voting} procedure that aggregates $\mL$ to reach the cross-model consensus of explanations, i.e., the \textit{consensus}, $\vc$ for $d_i$. 
	Specifically, the $k${-th} element of the consensus $\vc$ is $\vc_k = \frac{1}{m} \frac{\sum_{i=1}^m \mL_{ik}^2}{\| \mL_{i} \|},\ \forall 1\leq k\leq K$
	for LIME, where $K$ refers to the dimension of an explanation result and $\vc_k = \frac{1}{m} \sum_{i=1}^m \frac{\mL_{ik} - min(\mL_{i})}{max(\mL_{i}) - min(\mL_{i})},\ \forall 1\leq k\leq K$
	for SmoothGrad, following the conventional normalization-averaging procedure~\cite{ribeiro2016should,ahern2019normlime,smilkov2017smoothgrad}.
	To the end, the consensus has been reached for every sample in the target dataset based on committee voting.
	
	
      

	\textbf{Step3: Consensus-based Similarity Score.} Given the consensus, the approach calculates the \textit{consensus score} of every model in the committee by considering the similarity between the explanation result of each individual model and the consensus. 
	Specifically, for the explanations and the consensus based on LIME (visual feature importance in superpixel levels), cosine similarity between the flattened vector of explanation of each model and the consensus is used.
	For the results based on SmoothGrad (visual feature importance in pixel levels), a similar procedure is followed, where the proposed algorithm uses Radial Basis Function ($exp({-\frac{1}{2}(||\va - \vb||/\sigma)^2})$) for the similarity measurement. 
	The difference in similarity computations is due to that (1) the dimensions of LIME explanations are various for different samples while invariant for SmoothGrad explanations; (2) the scales of LIME explanation results vary much larger than SmoothGrad.
	Thus cosine similarity is more suitable for LIME while RBF is for SmoothGrad.
	Eventually, the framework computes a quantitative but relative score for each model in the committee using their similarity to the consensus.
	
	
	For further clarity, these three steps of the proposed framework are illustrated in Figure \ref{fig:pipeline} and formalized in Algorithm \ref{algo:consensus-short}, with more details in the appendix.

	\section{Overall Experiments and Results}
	\label{sec:overall-evaluation}
	
	In this section, we start by introducing the experiment setups.
	We use the image classification as the target task and follow the proposed framework to obtain the consensus and compute the consensus scores.
	Through the experiments, we have found (1) the alignment between the consensus and image semantic segmentation, (2) positive correlations between the consensus score and model performance, and (3) coincidental correlations between the consensus score and model interpretability.
	We end this section by robustness analyses of the framework.

	\subsection{Evaluation Setups}\label{sec:exprimental-details}

	
	\textbf{Datasets.}
	For overall evaluations and comparisons, we use ImageNet~\citep{deng2009imagenet} for general visual object recognition and CUB-200-2011~\citep{welinderetal2010caltech} for bird recognition respectively. 
	Note that ImageNet provides the class label for every image, and the CUB-200-2011 dataset includes the class label and pixel-level segmentation for the bird in every image, where the pixel annotations of visual objects are found to be aligned with the consensus.
	
	
	\textbf{Models.}
	For fair comparisons, we use more than 80 deep models trained on ImageNet that are publicly available\footnote{\url{https://github.com/PaddlePaddle/models/blob/release/1.8/PaddleCV/image_classification/README_en.md\#supported-models-and-performances}}. We also derive models on the CUB-200-2011 dataset through standard fine-tuning procedures. In our experiments, we include these models of the two committees based on ImageNet and CUB-200-2011 respectively. Both of them target at the \textit{image classification} task with each image being labeled to one category.
	
	
	\textbf{Interpretation Algorithms.}
	As we previously introduced, we consider two interpretation algorithms, LIME~\citep{ribeiro2016should} and SmoothGrad~\citep{smilkov2017smoothgrad}. Specifically, LIME surrogates the explanation as the assignment of visual feature importance to superpixels~\citep{vedaldi2008quick}, and SmoothGrad outputs the explanations as the visual feature importance over pixels. 
	In this way, we can validate the flexibility of the proposed framework over explanation results from diverse sources (i.e., linear surrogates vs. input gradients) and in multiple granularity (i.e., feature importance in superpixel/pixel-levels).

    \subsection{Alignment between the Consensus and Image Segmentation}
    
    \begin{figure*}[h]
		\centering
		\subfloat{\includegraphics[width=0.98\linewidth]{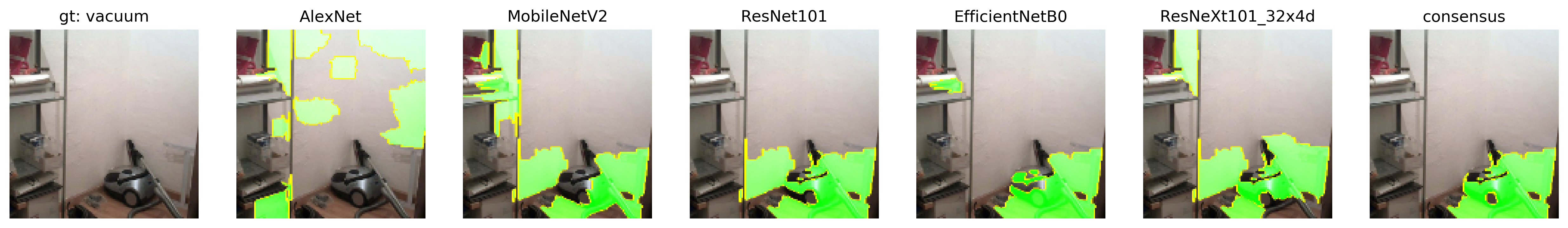}}
		
		\subfloat{\includegraphics[width=0.98\linewidth]{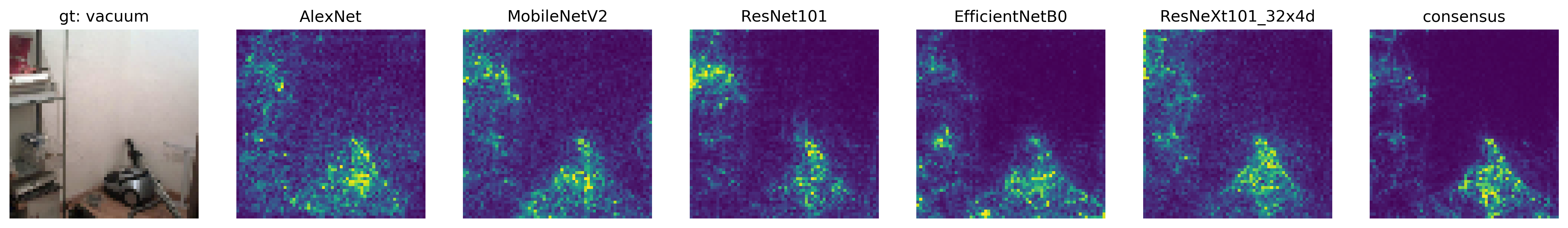}}
		\caption{Visual comparisons between consensus and the interpretation results of CNNs using LIME (in the upper line) and SmoothGrad (in the lower line) based on an image from ImageNet, where the ground truth of segmentation is not available.}
		\label{fig:imagenet_models_qualitative}
	\end{figure*}

	\begin{figure*}[h]
		\centering
		\subfloat{\includegraphics[width=0.98\linewidth]{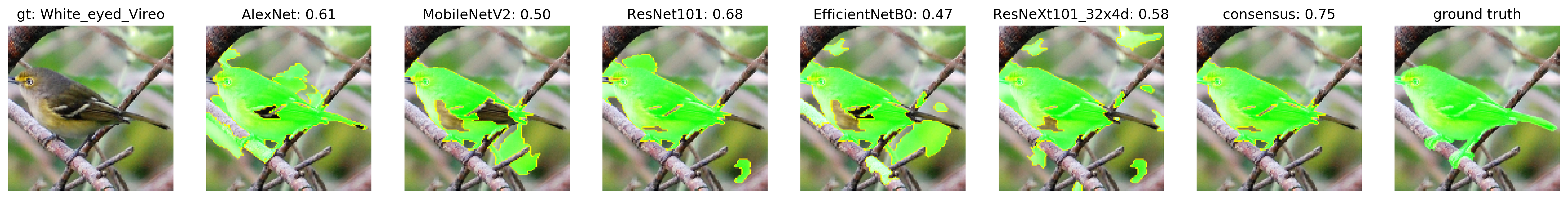}}
		
		\subfloat{\includegraphics[width=0.98\linewidth]{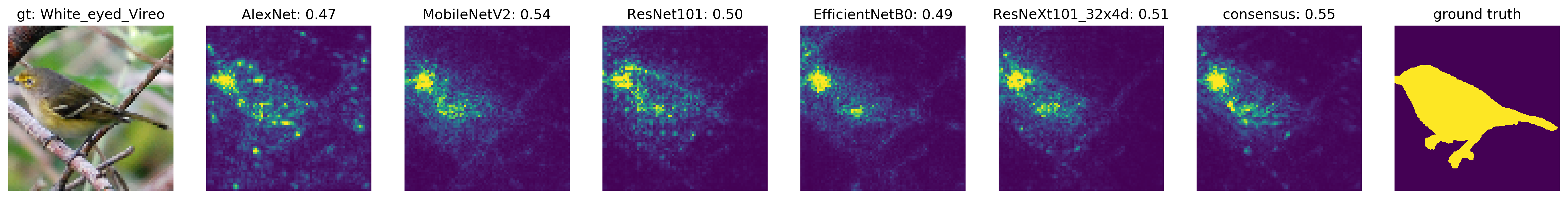}}
		\caption{Visual comparisons between consensus and the explanation results of deep models using LIME (in the upper line) and SmoothGrad (in the lower line) based on an image from CUB-200-2011, where the ground truth of segmentation is available as pixel-wise annotations and the mean Average Precision (mAP) are measured.}
		\label{fig:cub_models_qualitative}
	\end{figure*}
	
	The image segmentation task searches the pixel-wise classifications of images.
	Cross-model consensus of explanations for image classification are well aligned to image segmentation, especially when only one main object is contained in the image.
	This partially demonstrates the effectiveness of most deep models in extracting visual objects from input images.
    We show two examples using both LIME and SmoothGrad in Figure~\ref{fig:imagenet_models_qualitative} and \ref{fig:cub_models_qualitative} from ImageNet and CUB-200-2011 respectively.
    More examples can be found in appendix.
    
\begin{figure*}[t]
		\centering
		\subfloat[LIME]{\includegraphics[width=0.4\linewidth]{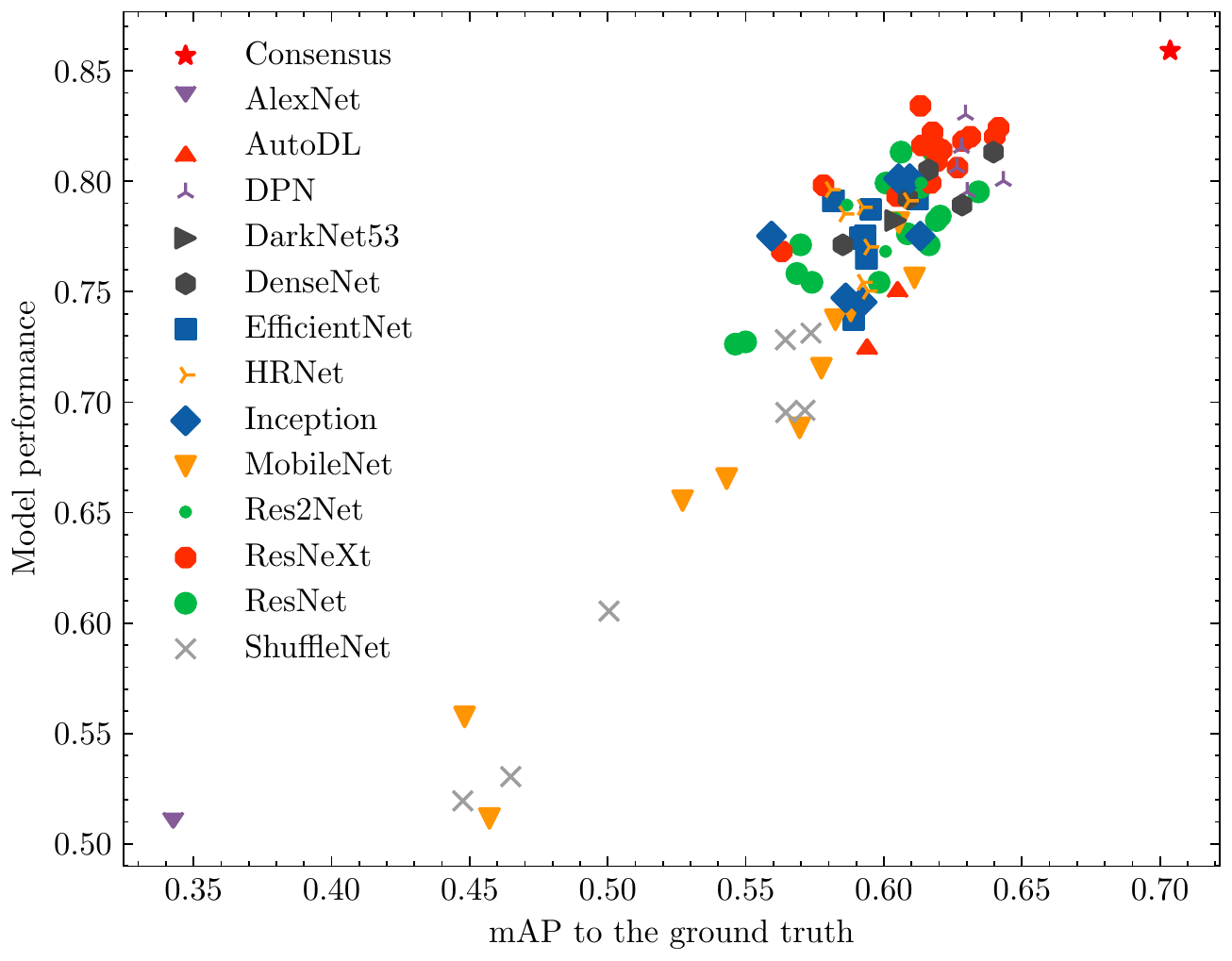}}
		\subfloat[SmoothGrad]{\includegraphics[width=0.4\linewidth]{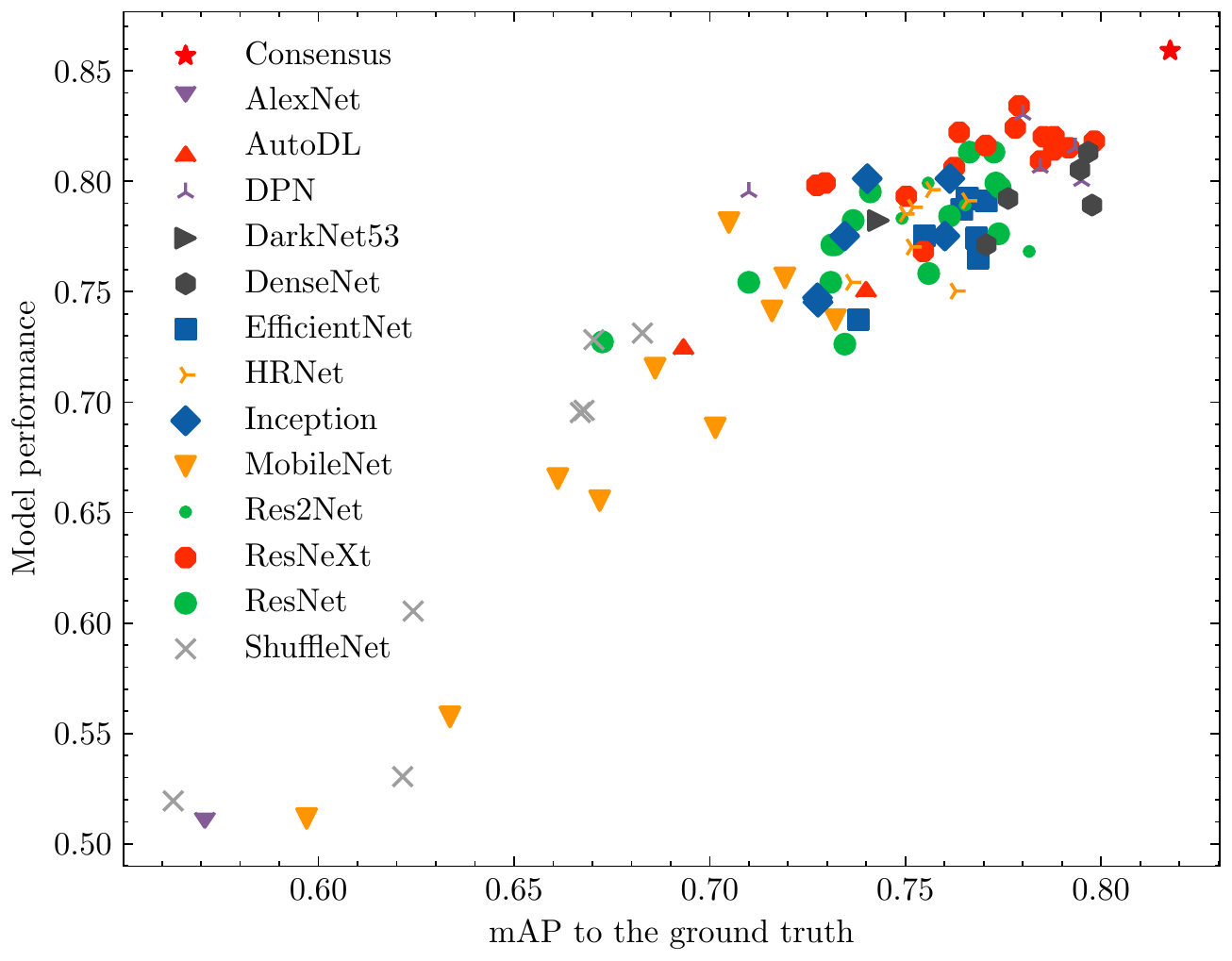}}
		\caption{Correlation between model performance and mAP to the segmentation ground truth using (a) LIME and (b) SmoothGrad with CUB-200-2011 over 85 models. 
		Pearson correlation coefficients are 0.927 (with p-value 4e-37) for LIME and 0.916 (with p-value 9e-35) for SmoothGrad.
		The points ``Consensus'' here refer to the testing accuracy of the ensemble of networks in the committee by probabilities averaging and voting (in y-axis), as well as the mAP between the consensus and the ground truth (in x-axis).}
		\label{fig:cub_all_models_map}
	\end{figure*}

    To quantitatively demonstrate the alignment, we compute the Average Precision (AP) score between the cross-model consensus of explanations and the image segmentation ground truth on CUB-200-2011, where the latter is available.
    We further take the mean of AP scores (mAP) over the dataset to compare with the overall consensus scores.
    Figure~\ref{fig:cub_all_models_map} shows the results, where the consensus achieves higher mAP scores than any individual network.
    Both quantitative results and visual comparisons in validate the closeness of consensus to the ground truth of image segmentation.
    
    
	\subsection{Positive Correlations between Consensus Scores and Model Performance}
	\label{sec:with-performance}
	
    \begin{figure*}[t]
        \centering
    	\subfloat[LIME on ImageNet]{\includegraphics[width=0.32\linewidth]{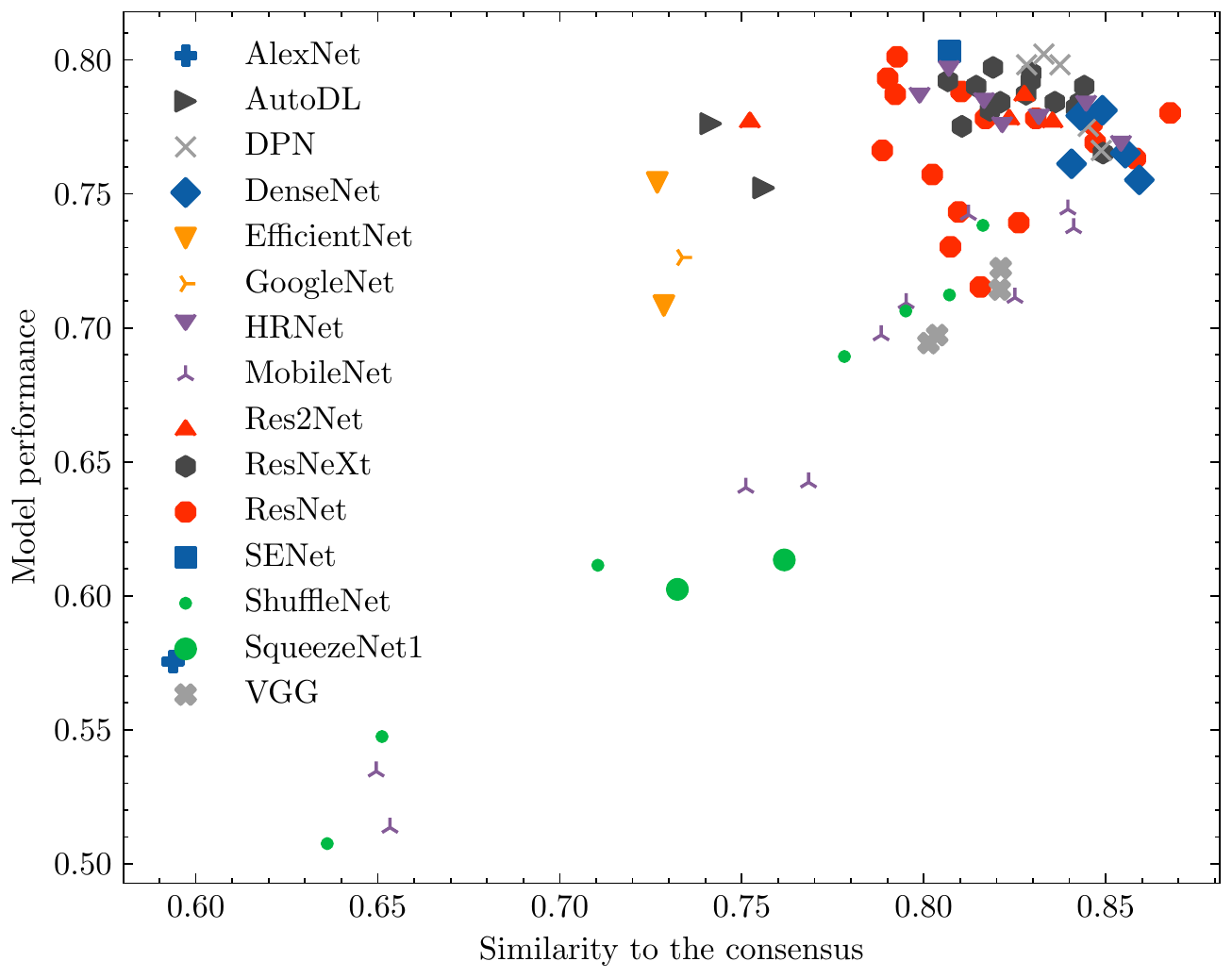}}
    	\subfloat[SmoothGrad on ImageNet]{\includegraphics[width=0.32\linewidth]{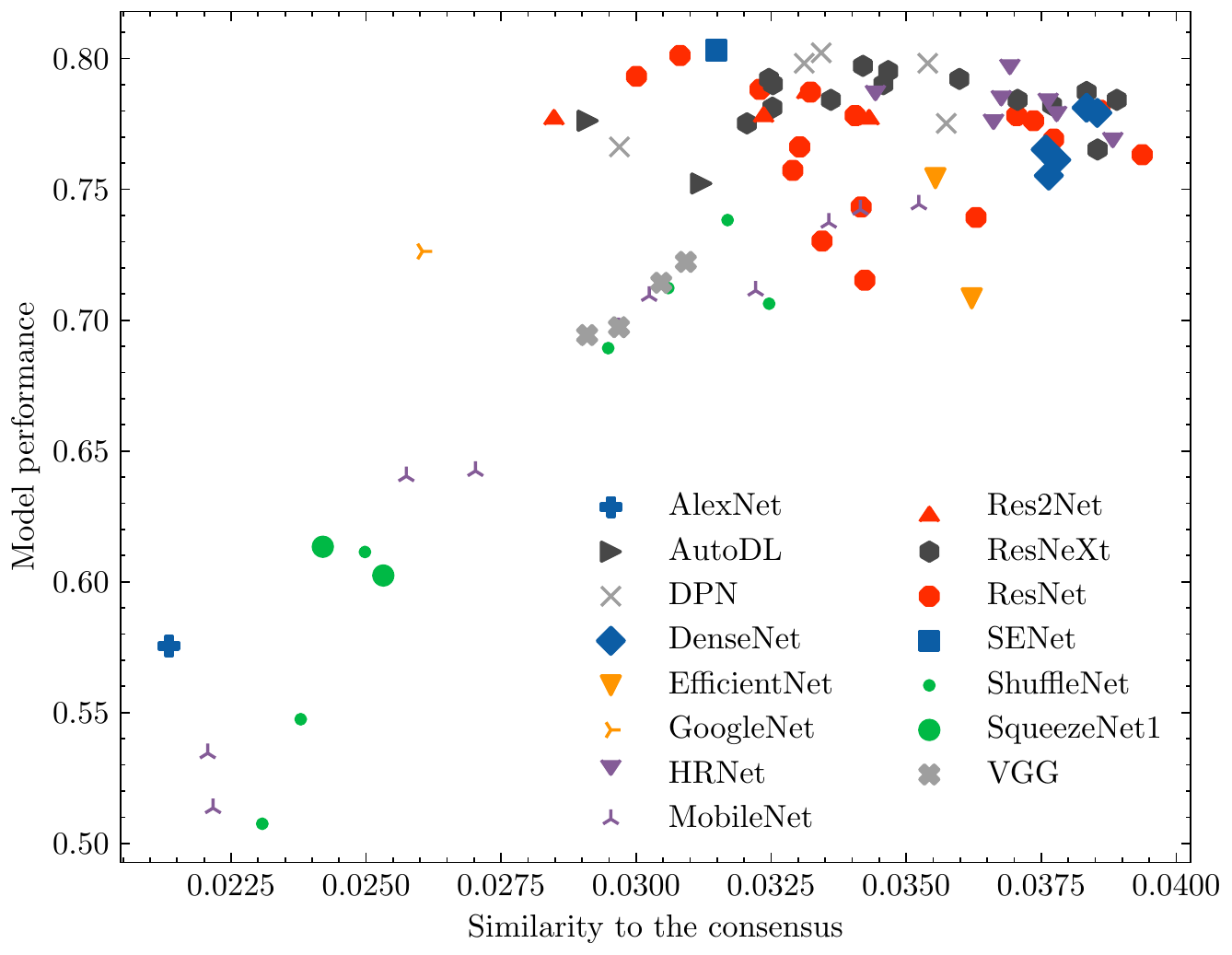}}
    	\subfloat[\scriptsize LIME vs. SmoothGrad on ImageNet]{\includegraphics[width=0.32\linewidth]{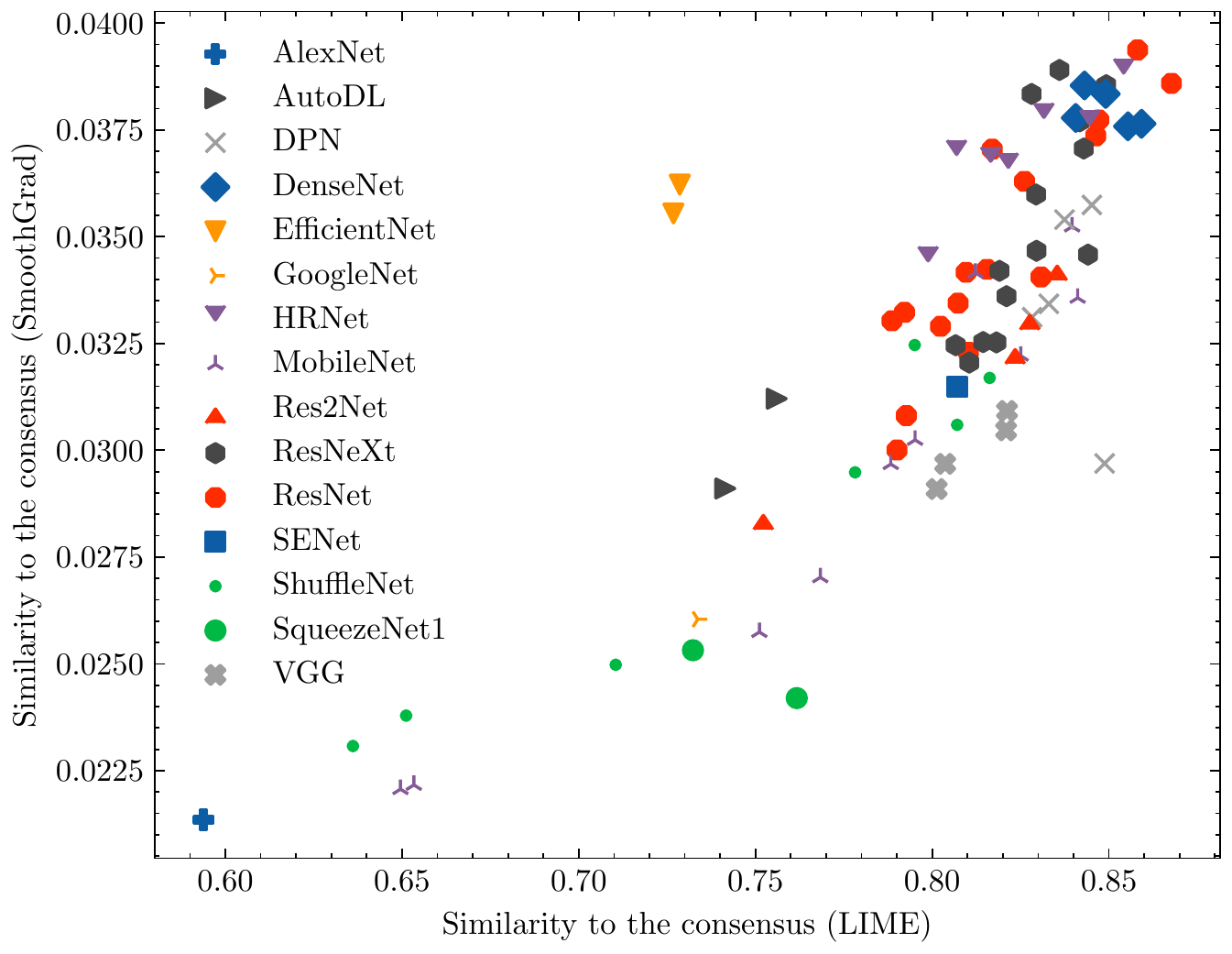}}
    	
    	\subfloat[LIME on CUB]{\includegraphics[width=0.32\linewidth]{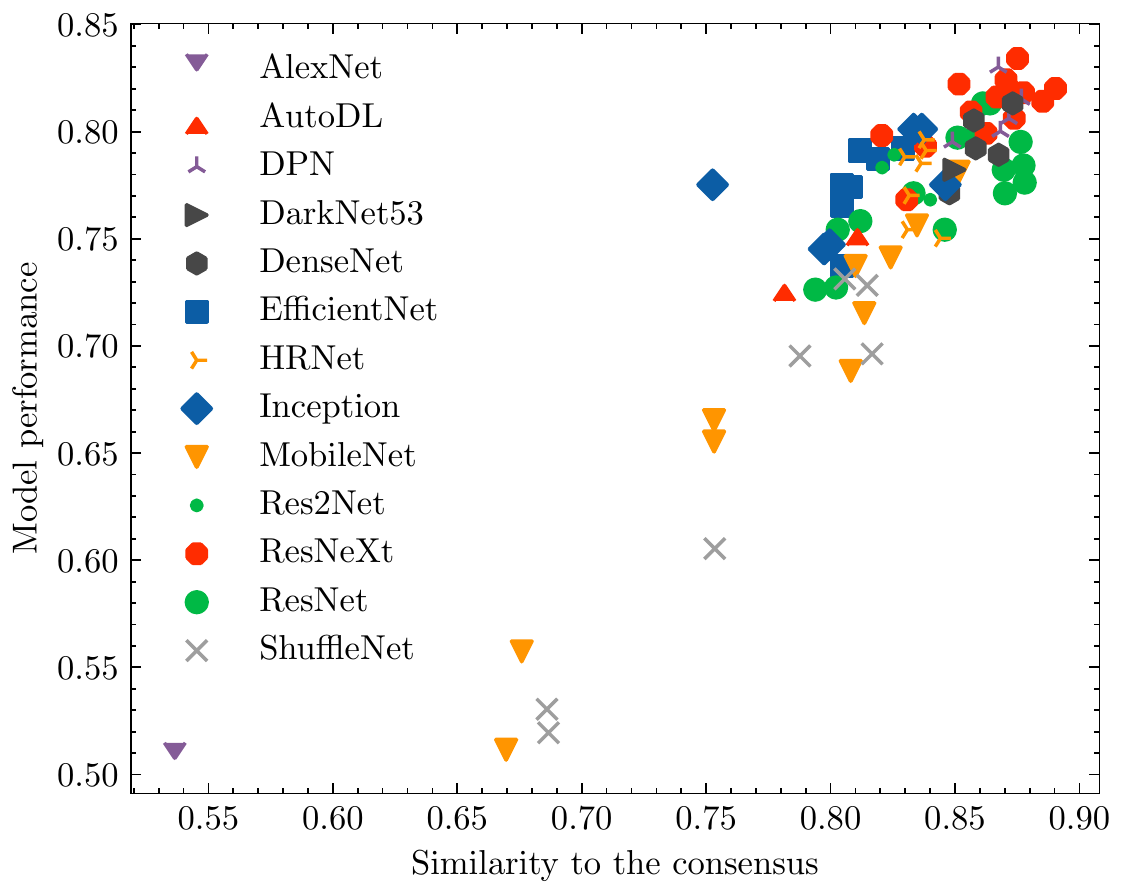}}
    	\subfloat[SmoothGrad on CUB]{\includegraphics[width=0.32\linewidth]{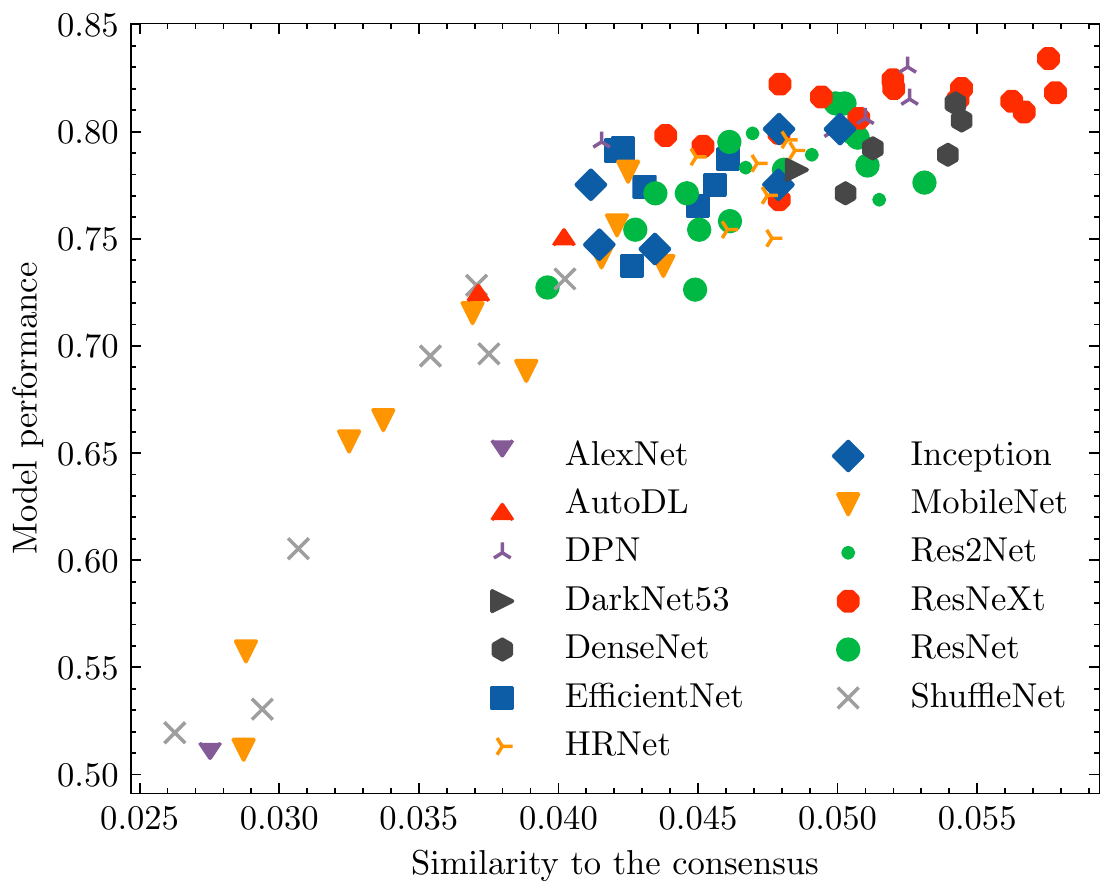}}
    	\subfloat[LIME vs. SmoothGrad on CUB]{\includegraphics[width=0.32\linewidth]{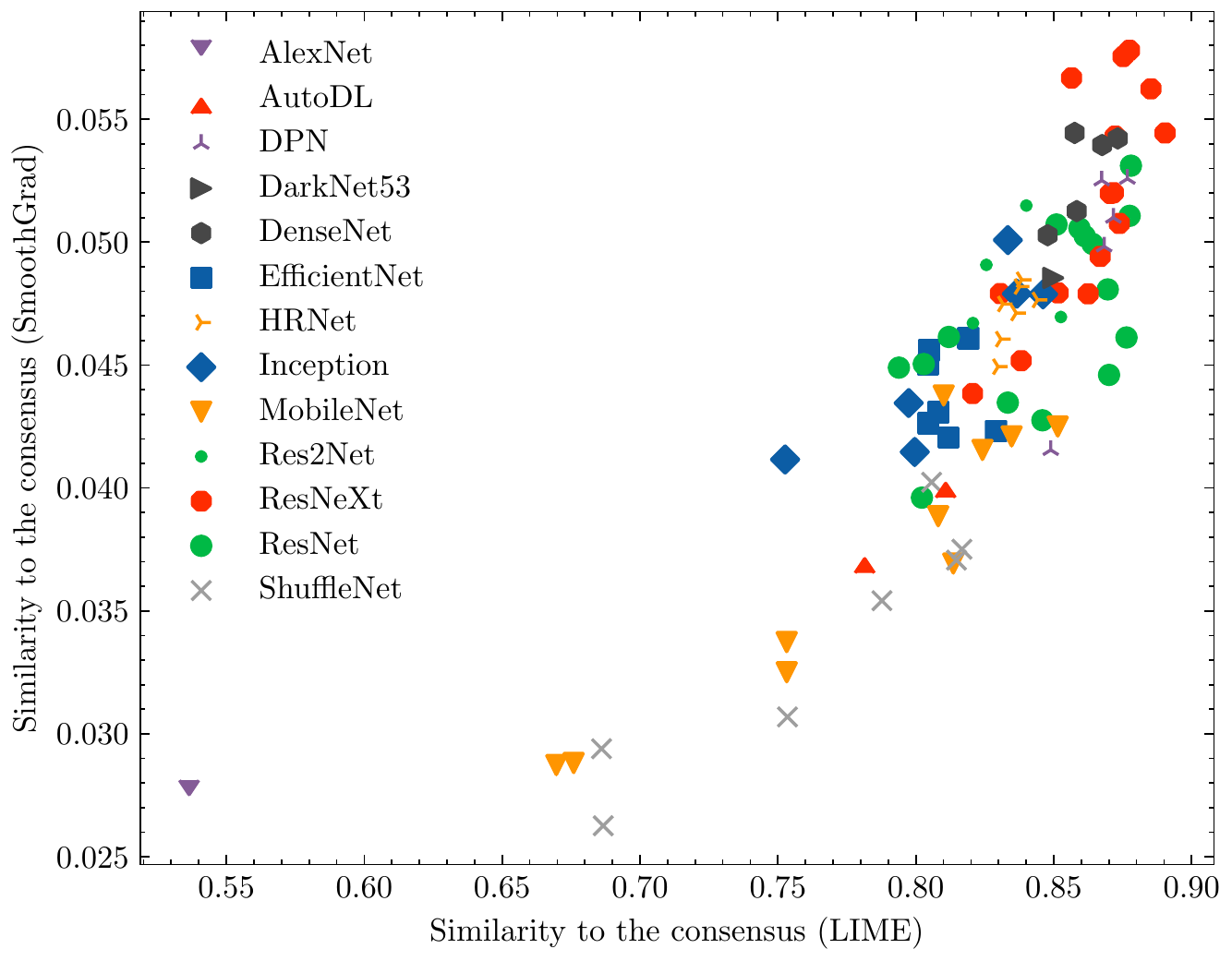}}
    	\caption{Model performance v.s. similarity to the consensus using LIME (a,d) and SmoothGrad (b,e) over 81 models on ImageNet (a,b) and 85 models on CUB-200-2011 (d,e). The third column shows the similarity to the consensus of SmoothGrad interpretations v.s. similarity to the consensus of LIME interpretations on ImageNet committee (c) and CUB-200-2011 committee (f). 
        Pearson correlation coefficients are (a) 0.8087, (b) 0.783, (c) 0.825, (d) 0.908, (e) 0.880 and (f) 0.854.
        For concise purpose, networks in the same family are represented by the same symbol.
    	}
    	\label{fig:all_results}
    \end{figure*}
	
	
    
    Figure~\ref{fig:all_results} shows the positive correlations between the similarity score to the consensus (in x-axis) and model performance (in y-axis).
    Specifically, in Figure~\ref{fig:all_results} (a-b) and (d-e), we present the results using LIME (a,d) and SmoothGrad (b,e) on ImageNet (a,b) and CUB-200-2011 (d,e).
    All correlations here are strong with significance tests passed, though in some local areas of the correlation plots between the consensus score and model performance.
    In this way, we could conclude that, in an overall manner, the evaluation results based on the consensus score using both LIME and SmoothGrad over the two datasets are correlated to model performance with significance.
    More experiments on other datasets with random subsets of deep models will be shown in in Figure \ref{fig:more_committees} (Section \ref{sec:robustness}).

\subsection{``Coincidental'' Correlations between Consensus Scores and Model Interpretability}
    \label{sec:comparison}
    
    Deep model interpretability measures \textit{the ability to present in understandable terms to a human}~\citep{doshi2017towards}.
    While no formal and agreed measurements for the interpretability evaluation, two evaluation methods, i.e., Network Dissection~\citep{bau2017network} and user-study experiments, are quite common for this purpose.
    Though the proposed framework and the consensus scores are based on explanation results, they do not directly estimate the model interpretability.
    Nevertheless, in this subsection, we present the coincidental correlations between the consensus scores and the interpretability measurements.
    
    
    \begin{table*}[t]
    \centering
    \caption{Rankings (and scores) of five deep models, evaluated by Network Dissection~\citep{bau2017network}, user-study evaluations, and the proposed framework with LIME and SmoothGrad. 
    }
    \begin{tabular}{@{}llllll@{}}
    \toprule
    \multicolumn{1}{r}{}                         & DenseNet161 & ResNet152 & VGG16 & GoogleNet & AlexNet \\ \midrule
    Network Dissection                           & 2           & 1         & 3     & 4         & 5       \\
    User-Study Evaluations                                 & 1 (1.715)          & 2 (1.625)        & 3 (1.585)    & 4 (1.170)        & 5 (0.840)   \\
    Consensus (LIME)       & 1 (0.849)          & 2 (0.846)     & 3 (0.821)    & 4 (0.734)        & 5 (0.594)      \\
    Consensus (SmoothGrad) & 1 (0.038)          & 2 (0.037)        & 3 (0.030)    & 4 (0.026)        & 5 (0.021)      \\ \bottomrule
    \end{tabular}
    \label{table:rankings}
    \end{table*}
    
    \textbf{Consensus \textit{versus} Network Dissection.}
    We compare the results of the proposed framework with the interpretability evaluation solution Network Dissection~\citep{bau2017network}.
	On the Broden dataset, Network Dissection reported a ranking list of five models (w.r.t. the model interpretability), shown in Table \ref{table:rankings}, through counting the semantic neurons, where a neuron is defined \textit{semantic} if its activated feature maps overlap with human-annotated visual concepts.
	Based on the proposed framework, we report the consensus scores of using LIME and SmoothGrad in Table \ref{table:rankings}, which are consistent to Figure \ref{fig:all_results} (a, LIME) and (b, SmoothGrad).
	The three ranking lists are almost identical, except the comparisons between DenseNet161 and ResNet152, where in the both lists based on the consensus score, DenseNet161 is similar to ResNet152 with marginally elevated consensus scores, while Network Dissection considers ResNet152 is more interpretable than DenseNet161. 
	
	We believe the results from our proposed framework and Network Dissection are close enough from the perspectives of ranking lists. 
	The difference may be caused by the different ways that our framework and Network Dissection perform the evaluations. 
	The consensus score measures the similarity to the consensus explanations on images, while Network Dissection counts the number of neurons in the intermediate layers activated by all the visual concepts, including objects, object parts, colors, materials, textures and scenes. 
	Furthermore, Network Dissection evaluates the interpretability of deep models using the Broden dataset with densely labeled visual objects and patterns \citep{bau2017network}, while the consensus score does not need additional datasets or the ground truth of semantics. 
	In this way, the results by our proposed framework and Network Dissection might be slightly different.
	
	\textbf{Consensus \textit{versus} User-Study Evaluations.}
	In order to further validate the effectiveness of the proposed framework, we have also conducted user-study experiments on these five models and report the results on the second row of Table \ref{table:rankings}. 
	See the appendix for the experimental settings of the user-study evaluations.
	This confirms that our proposed framework is capable of approximating the model interpretability.
	
	\subsection{Robustness Analyses of Consensus}\label{sec:robustness}
    In this subsection, we investigate several factors that might affect the evaluation results with consensus, including the use of basic interpretation algorithms (e.g., LIME and SmoothGrad), the size of committee, and the candidate pool for models in the committee.
    
    \textbf{Consistency between LIME and SmoothGrad.}
	Even though the granularity of explanation results from LIME and SmoothGrad are different, which causes mismatching in mAP scores to segmentation ground truth, the consensus scores based on the two algorithms are generally consistent. 
	The consistency has been confirmed by Figure \ref{fig:all_results} (c, f), where the overall results based on LIME is strongly correlated to SmoothGrad over all models on both datasets.
	This shows that the proposed framework can work well with a wide spectrum of basic interpretation algorithms.
	

	
    \textbf{Consistency of Cross-Committee Evaluations.}
    In real-word applications, the committee-based estimations and evaluations may make inconsistent results in a committee-by-committee manner. 
    In this work, we are interested in whether the consensus score estimations are consistent against the change of committee. 
    Given 16 ResNet models as the targets, we form 20 independent committees through combining the 16 ResNet models with 10--20 models randomly drawn from the rest of networks. 
    In each of these 20 independent committees, we compute the consensus scores of the 16 ResNet models. 
    We then estimate the Pearson correlation coefficients between any of these 20 results and the one in Figure~\ref{fig:all_results}~(a), where the mean correlation coefficient is 0.96 with the standard deviation 0.04. 
    Thus, we can say the consensus score evaluation would be consistent against randomly picked committees.
    
	

    \begin{wrapfigure}[18]{r}{0.6\linewidth}
		\centering
		\includegraphics[width=\linewidth]{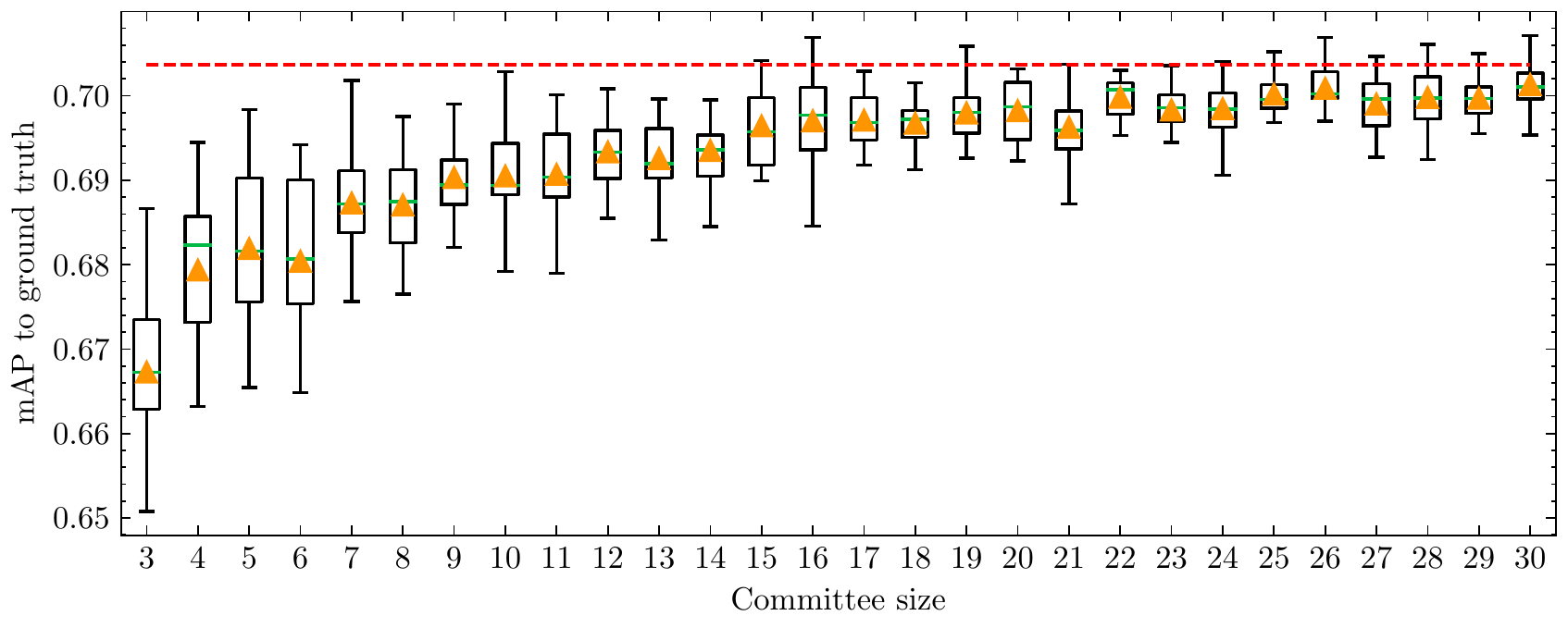}
		\caption{Convergence of mAP between the ground truth and the consensus results based on committees of increasing sizes, using LIME on CUB-200-2011. The green lines and orange triangles are, respectively, the mean values and the median values of 20 random trials. The red dashed line is the mAP of the consensus reached by the complete committee of the original 85 models.}
		\label{fig:cub_committee_size_map_lime}	
	\end{wrapfigure}

    \textbf{Convergence over Committee Sizes}
    To understand the effect of the committee size to the consensus score estimation, we run the proposed framework using committees of various sizes formed by deep models that are randomly picked up from the pools. In Figure~\ref{fig:cub_committee_size_map_lime}, we plot and compare the performance of the consensus with increasing committee sizes, where we estimate the mAP between the ground truth and the consensus reached by the random committees of different sizes and 20 random trials have been done for every single size independently. It shows that the curve of mAP would quickly converge to the complete committee, while the consensus based on a small proportion of committee (e.g., 15 networks) works good enough even compared to the complete committee of 85 networks.
	
	\textbf{Applicability with Random Committees over More Datasets.}
	To demonstrate the applicability of the proposed framework, we extend our experiments using networks randomly picked up from the pool to other datasets, including Stanford Cars 196 \citep{krause2013d}, Oxford Flowers 102 \citep{nilsback2008automated} and Foods 101 \citep{bossard2014food}.
	Dataset descriptions and experimental details are included in the appendix.
	The results in Figure \ref{fig:more_committees} confirm that the positive correlations between the consensus score and model performance exist for a wide range of models on ubiquitous datasets/tasks.
	
	\begin{figure*}[t]
	    \centering
		\subfloat{\includegraphics[width=0.325\linewidth]{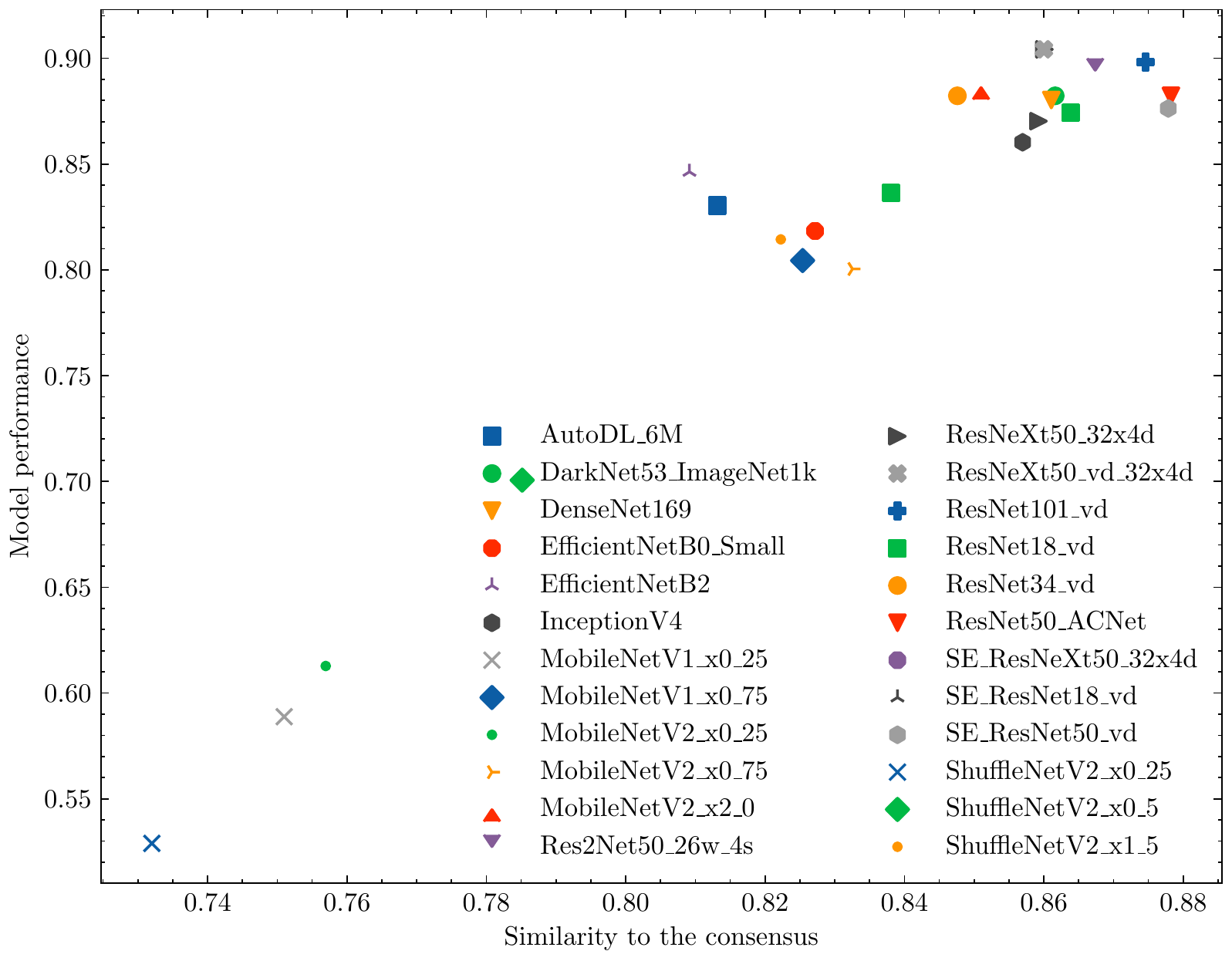}}
		\subfloat{\includegraphics[width=0.325\linewidth]{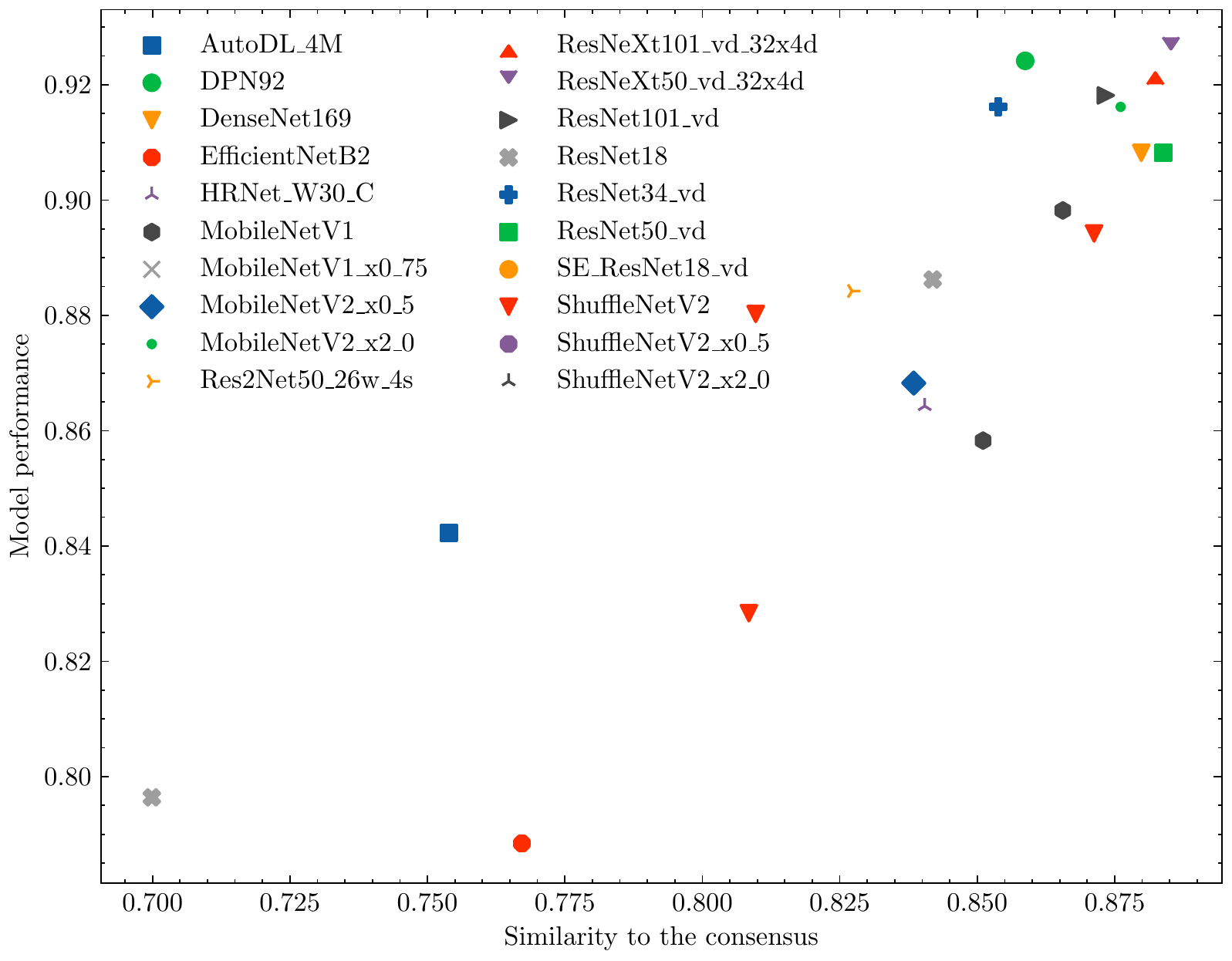}}
		\subfloat{\includegraphics[width=0.325\linewidth]{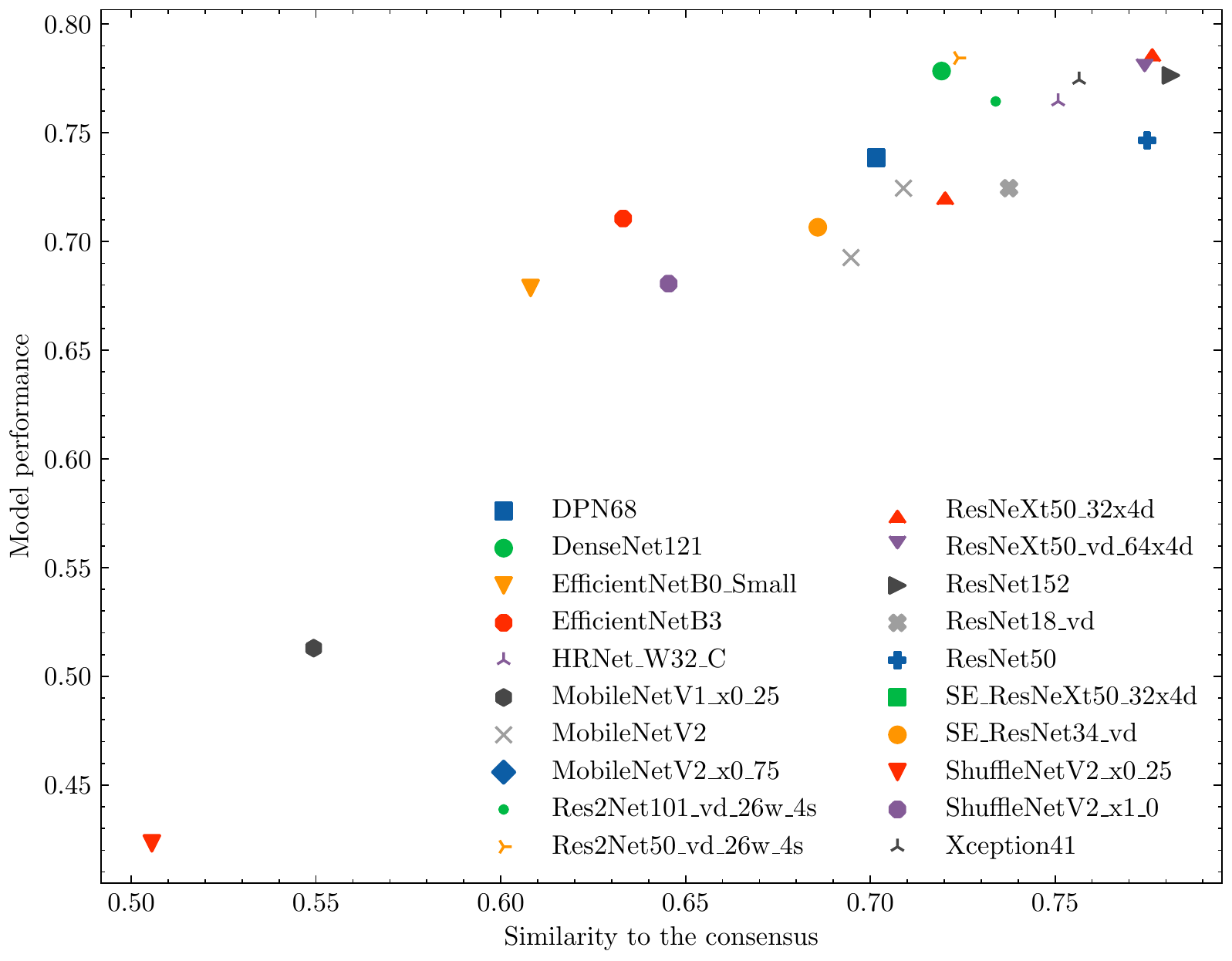}}
		\caption{Model performance v.s. the consensus scores using LIME on Stanford Cars 196 \citep{krause2013d}, Oxford Flowers 102 \citep{nilsback2008automated} and Foods 101 \citep{bossard2014food}. Pearson correlation coefficients are 0.9522, 0.8785 and 0.9134 respectively.}
		\label{fig:more_committees}
	\end{figure*}

    \section{Discussions: Limits and Potentials with Future Works}
    \label{sec:discussions-consensus}
    
    
    \textbf{Limits.} In this section, we would like to discuss several limits in our studies. First of all, we propose to study the features used by deep models for classification, but we use the explanation results (i.e., importance of superpixels/pixels in the image for prediction) obtained by interpretation algorithms. Obviously, the correctness of interpretation algorithms might affect our results. However, we use two independent algorithms, including  LIME~\cite{ribeiro2016should} and SmoothGrad~\cite{smilkov2017smoothgrad}, which attribute feature importance in two different scales i.e., superpixels and pixels. Both algorithms lead to the same observations and conclusive results (see Section~\ref{sec:robustness} for the consistency between results obtained by LIME and SmoothGrad). Thus, we believe the interpretation algorithms here are trustworthy and it is appropriate to use explanation results as a proxy to analyze features. For future research, we would include more advanced interpretation algorithms to confirm our observations.  
    
    We obtain some interesting observations from our experiments and make conclusions using multiple datasets. However, the image classification datasets used in our experiments have some limits --- every image in the dataset only consists of one visual object for classification. It is reasonable to doubt that when multiple visual objects (rather than the target for classification) and complicated visual patterns for background~\cite{koh2017understanding,chen2017targeted} co-exist in an image, the cross-model consensus of explanations may no longer overlap to the ground truth semantic segmentation. Actually, we include an example of COCO dataset~\cite{lin2014microsoft} in appendix, where multiple objects co-exist in the image and consensus may not always match the segmentation. Our future work would focus on the datasets with multiple visual objects and complicated background for object detection, segmentation, and multi-label classification tasks. 
    
    Finally, only well-known models with good performance have been included in the committee. It certainly would bring some bias in our analysis. However, in practice, these models would be one of the first choices or frequently used in many applications for relevance. In our future work, we would include more models with diverse performance to seek more observations.
    
    
    
     \textbf{Potentials.} In addition to the limits, our work also demonstrates several potentials of cross-model consensus of explanations for further studies. As was shown in Figure~\ref{fig:cub_committee_size_map_lime}, with a larger committee, the consensus would slowly converge to a stable set of common features that clearly aligns with the segmentation ground truth of the dataset. This experiment further demonstrates the capacity of consensus to precisely position the visual objects for classification. Thus, in our future work, we would like to use consensus based on a committee of image classification models to detect the position of visual objects in the image.
    
    
    
\begin{figure*}[t] 
		\centering
		\subfloat[Using LIME]{\includegraphics[width=0.4\linewidth]{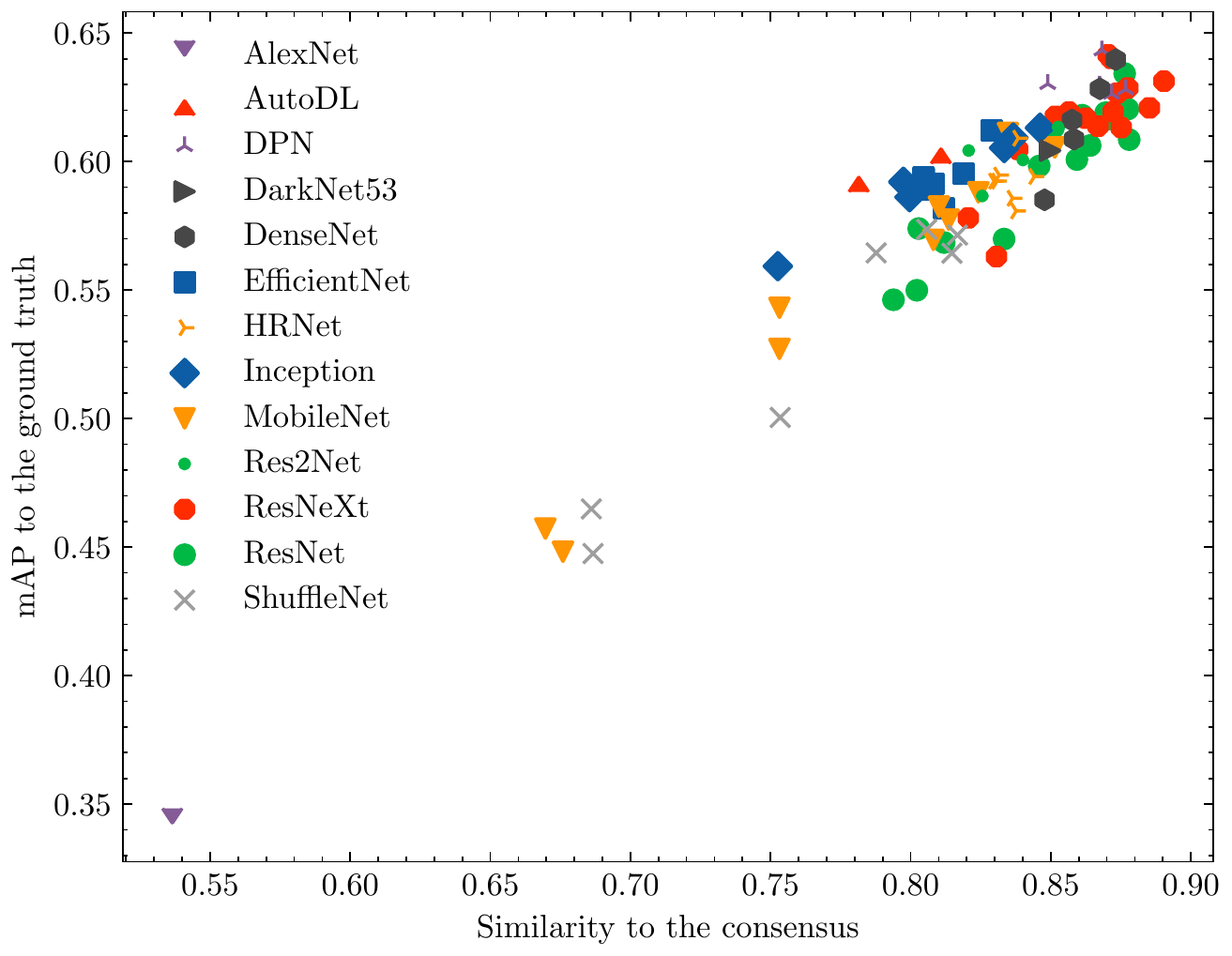}}
		\subfloat[Using SmoothGrad]{\includegraphics[width=0.4\linewidth]{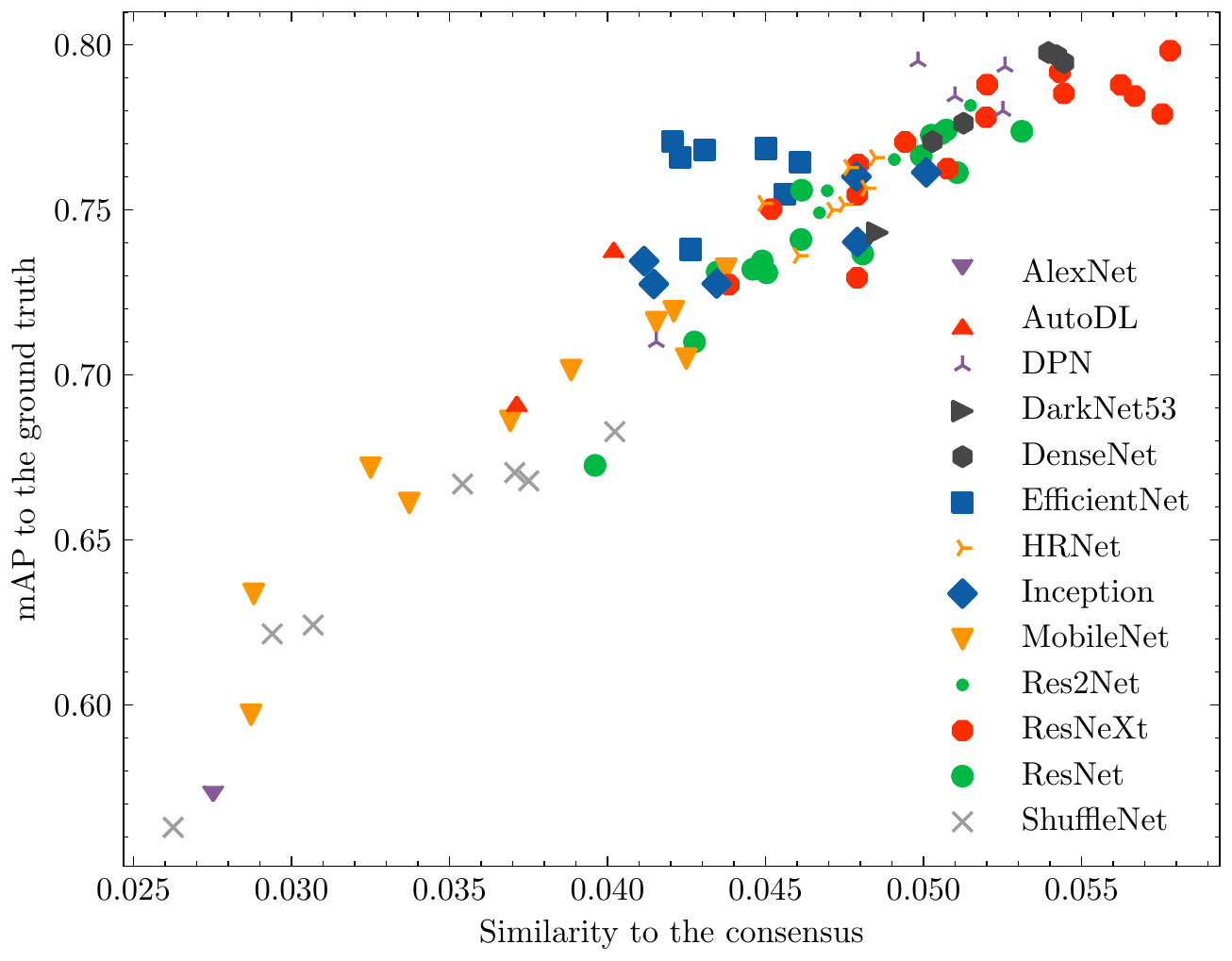}}
		\caption{Correlation between mAP scores to the segmentation ground truth and the consensus scores using (a) LIME and (b) SmoothGrad with the CUB-200-2011 dataset over 85 models (of the committee). 
		Pearson correlation coefficients are 0.885 (with p-value 3e-29) for LIME and 0.906 (with p-value 8e-33) for SmoothGrad.
        }
		\label{fig:cub_all_models_perf_map}
	\end{figure*}

    
    Furthermore, our experiments with both interpretation algorithms on all datasets have found that consensus scores are ``coincidentally'' correlated to the interpretability scores of the models, even though interpretability scores were evaluated through totally different ways --- network dissections~\cite{bau2017network} and user studies. Actually, network dissections evaluate the interpretability of a model through matching its activation maps in intermediate layers with the ground truth segmentation of visual concepts in the image. A model with higher interpretability should have more convolutional filters activated at the visual patterns/objects for classification. In this way, we particularly measure the similarity between the explanation results obtained for every model and the segmentation ground truth of images. We found that the models' segmentation-explanation similarity significantly correlates to their consensus score (see Figure~\ref{fig:cub_all_models_perf_map}). This observation might encourage us to further study the connections between interpretability and consensus scores in the future work.
    
    
    
	\section{Conclusion}
	
	In this paper, we study the common features shared by various deep models for image classification. We are wondering (1) what are the common features and (2) whether the use of common features could improve the performance. Specifically, given the explanation results obtained by interpretation algorithms, we propose to aggregate the explanation results from different models, and obtain the cross-model consensus of explanations through voting. To understand features used by every model and the common ones, we measure the consensus score as the similarity between the consensus and the explanation of every model. 
	
	Our empirical studies based on extensive experiments using 80+ deep models on 5 datasets/tasks find that (i) the consensus aligns with the ground truth semantic segmentation of the visual objects for classification; (ii) models with higher consensus scores would enjoy better testing accuracy; and (iii) the consensus scores coincidentally correlate to the interpretability scores obtained by network dissections and user evaluations. In addition to main claims, we also include additional experiments to demonstrate robustness of consensus, including the alternative use of LIME and SmoothGrad and their effects to the results/conclusions, consistency of consensus achieved by different groups of deep models, the fast convergence of consensus with increasing number of deep models in the committee, and random selection of deep models as the committee for consensus-based evaluation on the other datasets. All these studies confirm the applicability of consensus as a proxy to study and analyze the common features shared by different models in our research. Several open issues and potentials have been discussed with future directions introduced. Hereby, we are encouraged to further adopt consensus and consensus scores to understand the behaviors of deep models better.

\bibliography{reference}
\bibliographystyle{icml2021}

\newpage
\appendix




    \section{Complete Pseudocode of Cross-Model Consensus of Explanations}
    \label{sec:code}
    
    In the main text, Algorithm \ref{algo:consensus-short} presents the pseudocode of our framework of Cross-Model Consensus of Explanations.
    Here, Algorithm \ref{algo:consensus} completes the pseudocode of the framework with the details of the three functions that are used in Algorithm \ref{algo:consensus-short}.
    
    \begin{algorithm}[H]
	    \DontPrintSemicolon
        \caption{Functions in Algorithm~\ref{algo:consensus-short}.}
        
	    \label{algo:consensus}
	    
        \SetKwProg{Fn}{Function}{}{}
        
        \Fn{interpret ($\gA$, $d$, $\mM$):}{
            \tcc{ An interpretation algorithm $\gA$, a data sample $d$ and a trained model $\mM$. }
            
            \textbf{return} Explanation result of $\mM$ on $d$ by $\gA$.
        }
        
        \Fn{reach\_consensus($\mL$):}{
            \tcc{ $\mL$, a collection of interpretations of $m$ models for one given data sample. }
            
            \textbf{return} $\vc$, the consensus of the interpretations of $m$ models:
            $\vc_k = \frac{1}{m} \sum_{i=1}^m \frac{ \mL_{ik}^2 }{\| \mL_{i} \|}$ for LIME explanations;
            $\vc_k = \frac{1}{m} \sum_{i=1}^m \frac{\mL_{ik} - min(\mL_{i})}{max(\mL_{i}) - min(\mL_{i})}$ for SmoothGrad explanations.
        }
        
        \Fn{similarity($\va$, $\vb$):}{
            \tcc{ Two vectors $\va$ and $\vb$. }
            
            \textbf{return} The similarity score between $\va$ and $\vb$: $s = \frac{<\va,\ \vb>}{\lVert \va \rVert \lVert \vb \rVert}$ for LIME interpretations; $s = e^{-\frac{(\lVert \va - \vb \lVert /\sigma)^2}{2}}$ for SmoothGrad interpretations.
        }
        
    \end{algorithm}

    \section{Experimental Details}
    
    In this section, we present technique details of the experiments in the main text about the preparation of deep models for committee formations, the interpretation algorithms and the user-study evaluations.
    
    \subsection{Committee Formations}
    
    There are around 100 deep models trained on ImageNet that are publicly available\footnote{\url{https://github.com/PaddlePaddle/models/blob/release/1.8/PaddleCV/image_classification/README_en.md\#supported-models-and-performances}} at the moment we initiate the experiments.
	We first exclude some very large models that take much more computation resources.
	Then for the consistency of computing superpixels, we include only the models that take images of size 224$\times$224 as input, resulting 81 models for the committee based on ImageNet.
	For the intentions of comparing the models, a solution to including the models in the committee is to simply align the superpixels in different sizes of images.
	However, in our experiments, we choose not to do so since there are already a large number of available models.
	
	As for CUB-200-2011 \citep{welinderetal2010caltech}, similarly we first exclude the very large models.
	Then we follow the standard procedures \citep{sermanet2013overfeat,simonyan2015very} for fine-tuning ImageNet-pretrained models on CUB-200-2011.
	For simplicity, we use the same training setup for fine-tuning all pre-trained models on CUB-200-2011 (learning rate 0.01, batch size 64, SGD optimizer with momentum 0.9, resize to 256 being the short edge, randomly cropping images to the size of 224$\times$224), and obtain 85 models that are well trained.
	Different hyper-parameters may help to improve the performance of some specific networks, but for the same reason of the large number of available models, we choose not to search for better hyper-parameter settings.
	
	For Stanford Cars 196 \citep{krause2013d}, Oxford Flowers 102 \citep{nilsback2008automated} and Foods 101 \citep{bossard2014food}, we follow the same fine-tuning procedure as on CUB-200-2011.
	However, given the convergence over committee sizes (Figure \ref{fig:cub_committee_size_map_lime}), which suggests a committee of more than 15 models, we randomly choose around 20 models for each of the three datasets.
    
    \subsection{Interpretation Algorithms}
	
	To explain a deep model's predictions, LIME \citep{ribeiro2016should} on vision tasks first performs a superpixel segmentation \citep{vedaldi2008quick} for an image, then generates interpolated samples by randomly masking some superpixels and computing the outputs of the generated samples through the model, and finally fits the model outputs with the presence/absence of superpixels as input by a linear regression model.
	The linear weights then directly indicates the feature importance in superpixel level as the explanation result.
	
	The gradients of model output w.r.t. input can partly identify influential pixels, but due to the saturation of activation functions in the deep networks, the vanilla gradient is usually noisy.
	SmoothGrad \citep{smilkov2017smoothgrad} reduces the visual noise by repeatedly adding small random noises to the input so as to get a list of corresponding gradients, which are then averaged for the final explanation result.
	
	Note that many other interpretation algorithms are also available for our proposed framework while in this paper, we validate our approach with two trustworthy and commonly-used algorithms. 
    
	\subsection{Human User-Study Evaluations}
	
	As introduced in the main text, we have conducted user-study experiments for model interpretability over the five models that are discussed by Network Dissection, {i.e.}, DenseNet161, ResNet152, VGG16, GoogLeNet, AlexNet, and the evaluation result from user-study evaluations well aligns with results of our framework using either LIME or SmoothGrad.
	We describe here the experimental settings of the user-study evaluations.
	
	For each image, we randomly choose two models from the five models and present the LIME (or SmoothGrad respectively) explanations of the two models, without giving the model information to users.
	Users are then requested to choose which one helps better to reveal the model's reasoning of making predictions according to their understanding, or equal if the two interpretations are equally bad or good.
	Each pair of models is repeated three times and represented to different users.
	The better one in each pair will get three points and the other one will get zero; in the equal case, both get one point.
	Finally, a normalization of dividing the number of images and the number of repeats (i.e. 3) is performed for each model.
	The user-study evaluations yield the scores indicating the model interpretability, as shown in Table \ref{table:rankings}.
	
	\section{ResNet Family}
	
	\begin{figure*}[h]
		\centering
		\subfloat[Based on the Complete Committee]{\includegraphics[width=0.45\linewidth]{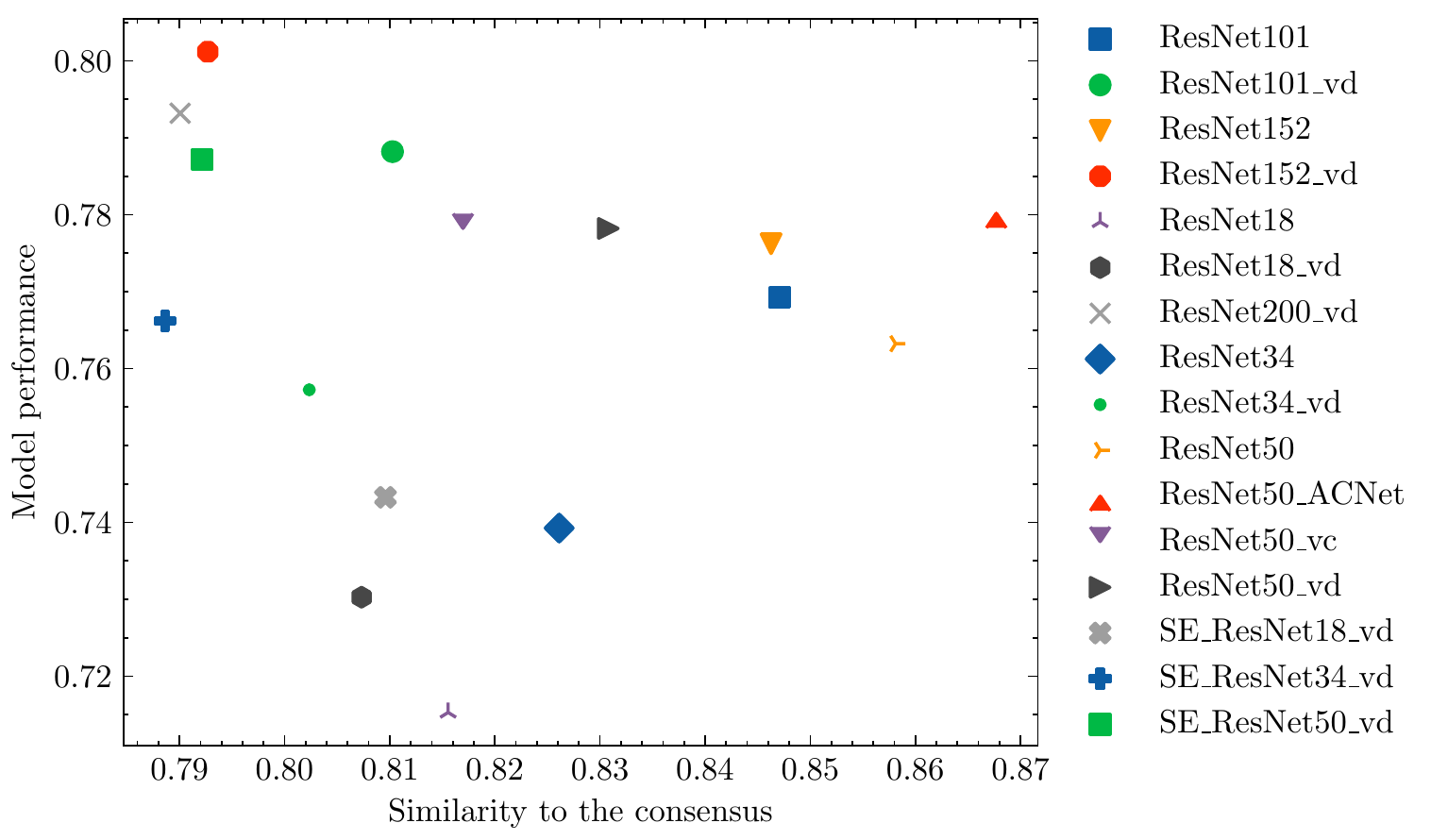}}
		\qquad
		\subfloat[Based on the ResNet Family]{\includegraphics[width=0.45\linewidth]{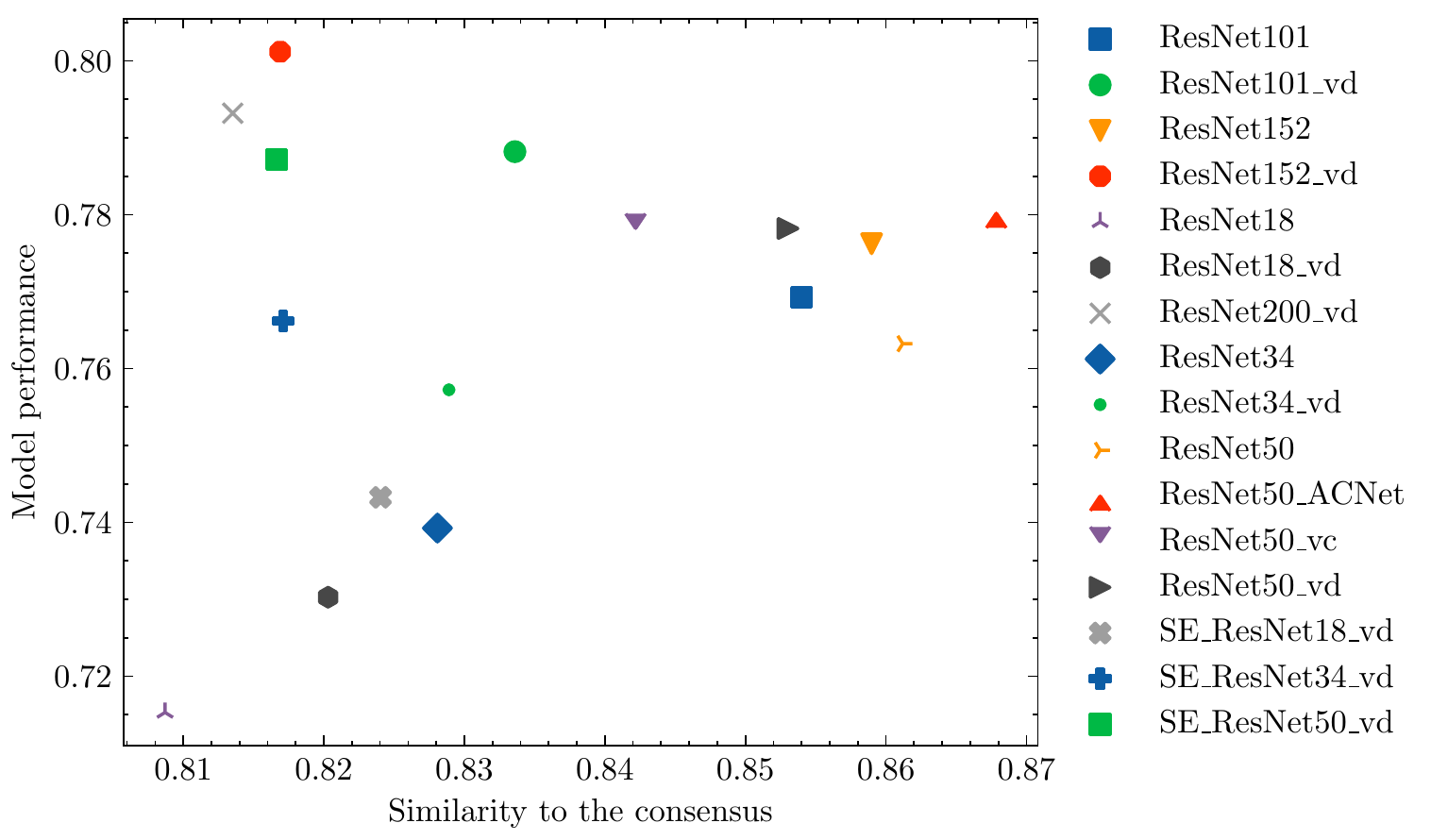}}
		\caption{Model performance v.s. similarity to the consensus of LIME on ResNet family. The consensus of (a) is voted by the complete committee on ImageNet (81 models), while the consensus of (b) is voted by ResNet family (16 models).}
		\label{fig:imagenet_resnet_models}
	\end{figure*}
	
	We show the zoomed plot of ResNet family (whose name contains ``ResNet'' key word) in the ImageNet-LIME committee of 81 models in Figure \ref{fig:imagenet_resnet_models} (a).
	Meanwhile, we also present the results using ResNet family as committee in Figure \ref{fig:imagenet_resnet_models} (b).
	These two subfigures have no large difference, which further confirms the consistency of our approach across different committees for ranking models.
	The positive correlation between the model performance and the consensus scores does not exist in the ResNet family, as we explained before that in some local areas, especially when models are extremely large, the correlation is not always positive. 
	
	

    \section{References of Network Structures}
	
    
    Most frequently-used structures of deep models have been evaluated in this paper, including AlexNet \citep{krizhevsky2012imagenet}, ResNet \citep{he2016deep}, ResNeXt \citep{xie2017aggregated}, SEResNet \citep{hu2018squeeze}, ShuffleNet \citep{zhang2018shufflenet,ma2018shufflenet}, MobileNet \citep{howard2017mobilenets,sandler2018mobilenetv2,howard2019searching}, VGG \citep{simonyan2015very}, GoogleNet \citep{szegedy2015going}, Inception \citep{szegedy2015going}, Xception \citep{chollet2017xception}, DarkNet \citep{redmon2016you,redmon2018yolov3}, DenseNet \citep{huang2017densely}, DPN \citep{chen2017dual}, SqueezeNet \citep{iandola2016squeezenet}, EfficientNet \citep{tan2019efficientnet}, Res2Net \citep{gao2019res2net}, HRNet \citep{wang2020deep}, Darts \citep{liu2018darts}, AcNet \citep{ding2019acnet} and their variants.
    
	
    \section{Numerical Report of Main Plots}
	
	Due to the large number of deep models evaluated, Figure~\ref{fig:cub_all_models_perf_map}, Figure~\ref{fig:cub_all_models_map} and Figure~\ref{fig:all_results} grouped some that are of the same architecture.
	Here, we report all of the corresponding numerical results in Table \ref{table:numbers} with a smaller scale.
	
	\begin{table*}[]
		\centering
		\caption{Numerical report of model performance and similarity to the consensus using LIME and SmoothGrad over 81 models on ImageNet in sub-table(a) corresponding to Figure \ref{fig:all_results} (a, b, c), and over 85 models on CUB-200-2011 in sub-table(b) corresponding to Figure \ref{fig:cub_all_models_map}, \ref{fig:cub_all_models_perf_map} and \ref{fig:all_results} (d, e, f). }
			\scalebox{0.5}{
			\subfloat[on ImageNet]{\begin{tabular}{@{}lrrr@{}}
				\toprule
				& \multicolumn{1}{l}{perf.} & \multicolumn{1}{l}{\begin{tabular}[c]{@{}l@{}}Consensus \\Scores w/ \\ LIME \end{tabular}} & \multicolumn{1}{l}{\begin{tabular}[c]{@{}l@{}}Consensus \\Scores w/ \\ Smooth-\\Grad \end{tabular}} \\ \midrule
				AlexNet &  0.575 &  0.594 &  0.0214 \\
				AutoDL\_4M &  0.752 &  0.756 &  0.0312 \\
				AutoDL\_6M &  0.776 &  0.741 &  0.0291 \\
				DPN107 &  0.798 &  0.828 &  0.0331 \\
				DPN131 &  0.802 &  0.833 &  0.0334 \\
				DPN68 &  0.766 &  0.849 &  0.0297 \\
				DPN92 &  0.775 &  0.845 &  0.0357 \\
				DPN98 &  0.798 &  0.837 &  0.0354 \\
				DenseNet121 &  0.755 &  0.859 &  0.0376 \\
				DenseNet161 &  0.781 &  0.849 &  0.0383 \\
				DenseNet169 &  0.765 &  0.855 &  0.0376 \\
				DenseNet201 &  0.779 &  0.843 &  0.0385 \\
				DenseNet264 &  0.761 &  0.841 &  0.0378 \\
				EfficientNetB0 &  0.754 &  0.727 &  0.0355 \\
				EfficientNetB0\_Small &  0.708 &  0.729 &  0.0362 \\
				GoogleNet &  0.726 &  0.734 &  0.0260 \\
				HRNet\_W18\_C &  0.766 &  0.854 &  0.0388 \\
				HRNet\_W30\_C &  0.776 &  0.832 &  0.0378 \\
				HRNet\_W32\_C &  0.781 &  0.845 &  0.0376 \\
				HRNet\_W40\_C &  0.773 &  0.822 &  0.0366 \\
				HRNet\_W44\_C &  0.782 &  0.817 &  0.0368 \\
				HRNet\_W48\_C &  0.794 &  0.807 &  0.0369 \\
				HRNet\_W64\_C &  0.784 &  0.799 &  0.0344 \\
				MobileNetV1 &  0.711 &  0.825 &  0.0322 \\
				MobileNetV1\_x0\_25 &  0.513 &  0.653 &  0.0222 \\
				MobileNetV1\_x0\_5 &  0.640 &  0.751 &  0.0257 \\
				MobileNetV1\_x0\_75 &  0.697 &  0.788 &  0.0297 \\
				MobileNetV2 &  0.742 &  0.812 &  0.0342 \\
				MobileNetV2\_x0\_25 &  0.534 &  0.650 &  0.0221 \\
				MobileNetV2\_x0\_5 &  0.642 &  0.768 &  0.0270 \\
				MobileNetV2\_x0\_75 &  0.709 &  0.795 &  0.0302 \\
				MobileNetV2\_x1\_5 &  0.737 &  0.841 &  0.0336 \\
				MobileNetV2\_x2\_0 &  0.744 &  0.840 &  0.0352 \\
				Res2Net101\_vd\_26w\_4s &  0.780 &  0.752 &  0.0285 \\
				Res2Net50\_14w\_8s &  0.781 &  0.823 &  0.0324 \\
				Res2Net50\_26w\_4s &  0.780 &  0.835 &  0.0343 \\
				Res2Net50\_vd\_26w\_4s &  0.790 &  0.828 &  0.0332 \\
				ResNeXt101\_32x4d &  0.784 &  0.843 &  0.0371 \\
				ResNeXt101\_vd\_32x4d &  0.795 &  0.830 &  0.0347 \\
				ResNeXt101\_vd\_64x4d &  0.784 &  0.821 &  0.0336 \\
				ResNeXt152\_32x4d &  0.782 &  0.842 &  0.0377 \\
				ResNeXt152\_64x4d &  0.787 &  0.828 &  0.0383 \\
				ResNeXt152\_vd\_32x4d &  0.792 &  0.807 &  0.0325 \\
				ResNeXt152\_vd\_64x4d &  0.790 &  0.814 &  0.0325 \\
				ResNeXt50\_32x4d &  0.765 &  0.849 &  0.0385 \\
				ResNeXt50\_64x4d &  0.784 &  0.836 &  0.0389 \\
				ResNeXt50\_vd\_32x4d &  0.790 &  0.844 &  0.0346 \\
				ResNeXt50\_vd\_64x4d &  0.792 &  0.829 &  0.0360 \\
				ResNet101 &  0.769 &  0.847 &  0.0377 \\
				ResNet101\_vd &  0.788 &  0.810 &  0.0323 \\
				ResNet152 &  0.776 &  0.846 &  0.0374 \\
				ResNet152\_vd &  0.801 &  0.793 &  0.0308 \\
				ResNet18 &  0.715 &  0.816 &  0.0342 \\
				ResNet18\_vd &  0.730 &  0.807 &  0.0334 \\
				ResNet200\_vd &  0.793 &  0.790 &  0.0300 \\
				ResNet34 &  0.739 &  0.826 &  0.0363 \\
				ResNet34\_vd &  0.757 &  0.802 &  0.0329 \\
				ResNet50 &  0.763 &  0.858 &  0.0394 \\
				ResNet50\_ACNet &  0.780 &  0.868 &  0.0386 \\
				ResNet50\_vc &  0.778 &  0.817 &  0.0370 \\
				ResNet50\_vd &  0.778 &  0.831 &  0.0341 \\
				SENet154\_vd &  0.803 &  0.807 &  0.0315 \\
				SE\_ResNeXt101\_32x4d &  0.781 &  0.818 &  0.0325 \\
				SE\_ResNeXt50\_32x4d &  0.775 &  0.810 &  0.0321 \\
				SE\_ResNeXt50\_vd\_32x4d &  0.797 &  0.819 &  0.0342 \\
				SE\_ResNet18\_vd &  0.743 &  0.810 &  0.0342 \\
				SE\_ResNet34\_vd &  0.766 &  0.789 &  0.0330 \\
				SE\_ResNet50\_vd &  0.787 &  0.792 &  0.0332 \\
				ShuffleNetV2 &  0.706 &  0.795 &  0.0325 \\
				ShuffleNetV2\_x0\_25 &  0.507 &  0.636 &  0.0231 \\
				ShuffleNetV2\_x0\_33 &  0.547 &  0.651 &  0.0238 \\
				ShuffleNetV2\_x0\_5 &  0.611 &  0.710 &  0.0250 \\
				ShuffleNetV2\_x1\_0 &  0.689 &  0.778 &  0.0295 \\
				ShuffleNetV2\_x1\_5 &  0.712 &  0.807 &  0.0306 \\
				ShuffleNetV2\_x2\_0 &  0.738 &  0.816 &  0.0317 \\
				SqueezeNet1\_0 &  0.602 &  0.732 &  0.0253 \\
				SqueezeNet1\_1 &  0.613 &  0.762 &  0.0242 \\
				VGG11 &  0.694 &  0.801 &  0.0291 \\
				VGG13 &  0.697 &  0.804 &  0.0297 \\
				VGG16 &  0.714 &  0.821 &  0.0305 \\
				VGG19 &  0.722 &  0.821 &  0.0309 \\ \bottomrule
			\end{tabular}
		}
		}
	\quad
		\scalebox{0.5}{
			\subfloat[on CUB-200-2011]{
				\begin{tabular}{@{}lrrrrr@{}}
					\toprule
					& \multicolumn{1}{l}{perf.} & \multicolumn{1}{l}{\begin{tabular}[c]{@{}l@{}}Consensus \\Scores w/ \\ LIME \end{tabular}} & \multicolumn{1}{l}{\begin{tabular}[c]{@{}l@{}}Consensus \\Scores w/ \\ Smooth-\\ Grad \end{tabular}} & \begin{tabular}[c]{@{}l@{}}mAP between\\ g.t. of \\ segmentation \\and LIME\\ explanation\end{tabular} & \begin{tabular}[c]{@{}l@{}}mAP between\\ g.t. of \\ segmentation \\and \\SmoothGrad\\ explanation\end{tabular} \\ \midrule
				AlexNet &  0.507 &  0.536 &  0.0275 &  0.343 &  0.571 \\
				AutoDL\_4M &  0.728 &  0.781 &  0.0371 &  0.594 &  0.693 \\
				AutoDL\_6M &  0.754 &  0.811 &  0.0402 &  0.605 &  0.740 \\
				DPN107 &  0.830 &  0.867 &  0.0525 &  0.630 &  0.780 \\
				DPN131 &  0.800 &  0.868 &  0.0498 &  0.643 &  0.795 \\
				DPN68 &  0.795 &  0.849 &  0.0415 &  0.630 &  0.710 \\
				DPN92 &  0.806 &  0.872 &  0.0510 &  0.626 &  0.784 \\
				DPN98 &  0.815 &  0.877 &  0.0526 &  0.628 &  0.793 \\
				DarkNet53\_ImageNet1k &  0.782 &  0.850 &  0.0485 &  0.604 &  0.743 \\
				DenseNet121 &  0.771 &  0.848 &  0.0503 &  0.585 &  0.771 \\
				DenseNet161 &  0.813 &  0.873 &  0.0542 &  0.640 &  0.797 \\
				DenseNet169 &  0.792 &  0.858 &  0.0513 &  0.609 &  0.776 \\
				DenseNet201 &  0.805 &  0.858 &  0.0544 &  0.616 &  0.795 \\
				DenseNet264 &  0.789 &  0.868 &  0.0540 &  0.628 &  0.798 \\
				EfficientNetB0 &  0.765 &  0.805 &  0.0450 &  0.594 &  0.769 \\
				EfficientNetB0\_Small &  0.737 &  0.805 &  0.0426 &  0.589 &  0.738 \\
				EfficientNetB1 &  0.775 &  0.805 &  0.0456 &  0.593 &  0.755 \\
				EfficientNetB2 &  0.787 &  0.819 &  0.0461 &  0.595 &  0.764 \\
				EfficientNetB3 &  0.791 &  0.812 &  0.0421 &  0.582 &  0.771 \\
				EfficientNetB4 &  0.792 &  0.829 &  0.0423 &  0.612 &  0.766 \\
				EfficientNetB5 &  0.774 &  0.808 &  0.0431 &  0.591 &  0.768 \\
				HRNet\_W18\_C &  0.754 &  0.831 &  0.0461 &  0.592 &  0.736 \\
				HRNet\_W30\_C &  0.770 &  0.832 &  0.0475 &  0.595 &  0.752 \\
				HRNet\_W32\_C &  0.785 &  0.836 &  0.0471 &  0.586 &  0.750 \\
				HRNet\_W40\_C &  0.750 &  0.844 &  0.0476 &  0.594 &  0.763 \\
				HRNet\_W44\_C &  0.788 &  0.830 &  0.0449 &  0.592 &  0.752 \\
				HRNet\_W48\_C &  0.796 &  0.838 &  0.0482 &  0.581 &  0.757 \\
				HRNet\_W64\_C &  0.791 &  0.838 &  0.0485 &  0.609 &  0.766 \\
				InceptionV4 &  0.745 &  0.797 &  0.0435 &  0.592 &  0.728 \\
				MobileNetV1 &  0.741 &  0.824 &  0.0415 &  0.588 &  0.716 \\
				MobileNetV1\_x0\_25 &  0.557 &  0.676 &  0.0288 &  0.448 &  0.634 \\
				MobileNetV1\_x0\_5 &  0.655 &  0.753 &  0.0325 &  0.527 &  0.672 \\
				MobileNetV1\_x0\_75 &  0.688 &  0.808 &  0.0388 &  0.569 &  0.701 \\
				MobileNetV2 &  0.737 &  0.810 &  0.0438 &  0.582 &  0.732 \\
				MobileNetV2\_x0\_25 &  0.511 &  0.670 &  0.0287 &  0.457 &  0.597 \\
				MobileNetV2\_x0\_5 &  0.665 &  0.753 &  0.0337 &  0.543 &  0.661 \\
				MobileNetV2\_x0\_75 &  0.715 &  0.814 &  0.0369 &  0.577 &  0.686 \\
				MobileNetV2\_x1\_5 &  0.756 &  0.835 &  0.0421 &  0.611 &  0.719 \\
				MobileNetV2\_x2\_0 &  0.781 &  0.851 &  0.0425 &  0.605 &  0.705 \\
				Res2Net101\_vd\_26w\_4s &  0.799 &  0.853 &  0.0470 &  0.613 &  0.756 \\
				Res2Net50\_14w\_8s &  0.789 &  0.826 &  0.0491 &  0.587 &  0.765 \\
				Res2Net50\_26w\_4s &  0.768 &  0.840 &  0.0515 &  0.601 &  0.782 \\
				Res2Net50\_vd\_26w\_4s &  0.783 &  0.821 &  0.0467 &  0.604 &  0.749 \\
				ResNeXt101\_32x4d &  0.818 &  0.877 &  0.0578 &  0.629 &  0.798 \\
				ResNeXt101\_32x8d\_wsl &  0.768 &  0.831 &  0.0479 &  0.563 &  0.755 \\
				ResNeXt101\_vd\_32x4d &  0.816 &  0.867 &  0.0494 &  0.614 &  0.771 \\
				ResNeXt101\_vd\_64x4d &  0.824 &  0.871 &  0.0520 &  0.642 &  0.778 \\
				ResNeXt152\_32x4d &  0.815 &  0.872 &  0.0543 &  0.619 &  0.792 \\
				ResNeXt152\_64x4d &  0.834 &  0.875 &  0.0576 &  0.613 &  0.779 \\
				ResNeXt152\_vd\_32x4d &  0.820 &  0.872 &  0.0520 &  0.640 &  0.788 \\
				ResNeXt152\_vd\_64x4d &  0.822 &  0.852 &  0.0479 &  0.618 &  0.764 \\
				ResNeXt50\_32x4d &  0.809 &  0.856 &  0.0567 &  0.619 &  0.785 \\
				ResNeXt50\_64x4d &  0.814 &  0.885 &  0.0562 &  0.621 &  0.788 \\
				ResNeXt50\_vd\_32x4d &  0.806 &  0.874 &  0.0508 &  0.627 &  0.762 \\
				ResNeXt50\_vd\_64x4d &  0.820 &  0.890 &  0.0544 &  0.631 &  0.785 \\
				ResNet101 &  0.784 &  0.878 &  0.0511 &  0.620 &  0.761 \\
				ResNet101\_vd &  0.813 &  0.864 &  0.0499 &  0.606 &  0.766 \\
				ResNet152 &  0.799 &  0.859 &  0.0506 &  0.601 &  0.773 \\
				ResNet152\_vd &  0.797 &  0.851 &  0.0507 &  0.613 &  0.774 \\
				ResNet18 &  0.726 &  0.794 &  0.0449 &  0.546 &  0.735 \\
				ResNet18\_vd &  0.754 &  0.846 &  0.0428 &  0.598 &  0.710 \\
				ResNet200\_vd &  0.813 &  0.861 &  0.0502 &  0.618 &  0.773 \\
				ResNet34 &  0.758 &  0.812 &  0.0461 &  0.569 &  0.756 \\
				ResNet34\_vd &  0.771 &  0.833 &  0.0435 &  0.570 &  0.731 \\
				ResNet50 &  0.776 &  0.878 &  0.0531 &  0.609 &  0.774 \\
				ResNet50\_ACNet &  0.782 &  0.870 &  0.0481 &  0.619 &  0.737 \\
				ResNet50\_vd &  0.795 &  0.876 &  0.0461 &  0.634 &  0.741 \\
				SE\_ResNeXt101\_32x4d &  0.793 &  0.838 &  0.0452 &  0.605 &  0.750 \\
				SE\_ResNeXt50\_32x4d &  0.798 &  0.821 &  0.0438 &  0.578 &  0.727 \\
				SE\_ResNeXt50\_vd\_32x4d &  0.799 &  0.863 &  0.0479 &  0.617 &  0.729 \\
				SE\_ResNet18\_vd &  0.727 &  0.802 &  0.0396 &  0.550 &  0.673 \\
				SE\_ResNet34\_vd &  0.754 &  0.803 &  0.0450 &  0.574 &  0.731 \\
				SE\_ResNet50\_vd &  0.771 &  0.870 &  0.0446 &  0.616 &  0.732 \\
				ShuffleNetV2 &  0.696 &  0.817 &  0.0375 &  0.571 &  0.668 \\
				ShuffleNetV2\_x0\_25 &  0.519 &  0.687 &  0.0263 &  0.448 &  0.563 \\
				ShuffleNetV2\_x0\_33 &  0.530 &  0.686 &  0.0294 &  0.465 &  0.622 \\
				ShuffleNetV2\_x0\_5 &  0.605 &  0.753 &  0.0307 &  0.500 &  0.624 \\
				ShuffleNetV2\_x1\_0 &  0.695 &  0.788 &  0.0354 &  0.564 &  0.667 \\
				ShuffleNetV2\_x1\_5 &  0.728 &  0.815 &  0.0371 &  0.564 &  0.670 \\
				ShuffleNetV2\_x2\_0 &  0.731 &  0.806 &  0.0402 &  0.574 &  0.683 \\
				Xception41 &  0.801 &  0.833 &  0.0501 &  0.605 &  0.761 \\
				Xception41\_deeplab &  0.775 &  0.753 &  0.0412 &  0.559 &  0.734 \\
				Xception65 &  0.801 &  0.837 &  0.0479 &  0.609 &  0.740 \\
				Xception65\_deeplab &  0.747 &  0.800 &  0.0415 &  0.586 &  0.728 \\
				Xception71 &  0.775 &  0.846 &  0.0479 &  0.613 &  0.760 \\ 
				consensus &  0.859 & N/A & N/A &  0.704 &  0.818 \\\bottomrule
			\end{tabular}
		}
		}
	
	\label{table:numbers}
	\end{table*}
    
    \section{More Visualization Results}
	
\begin{figure*}[t]
	    \centering
	    \includegraphics[width=0.8\linewidth]{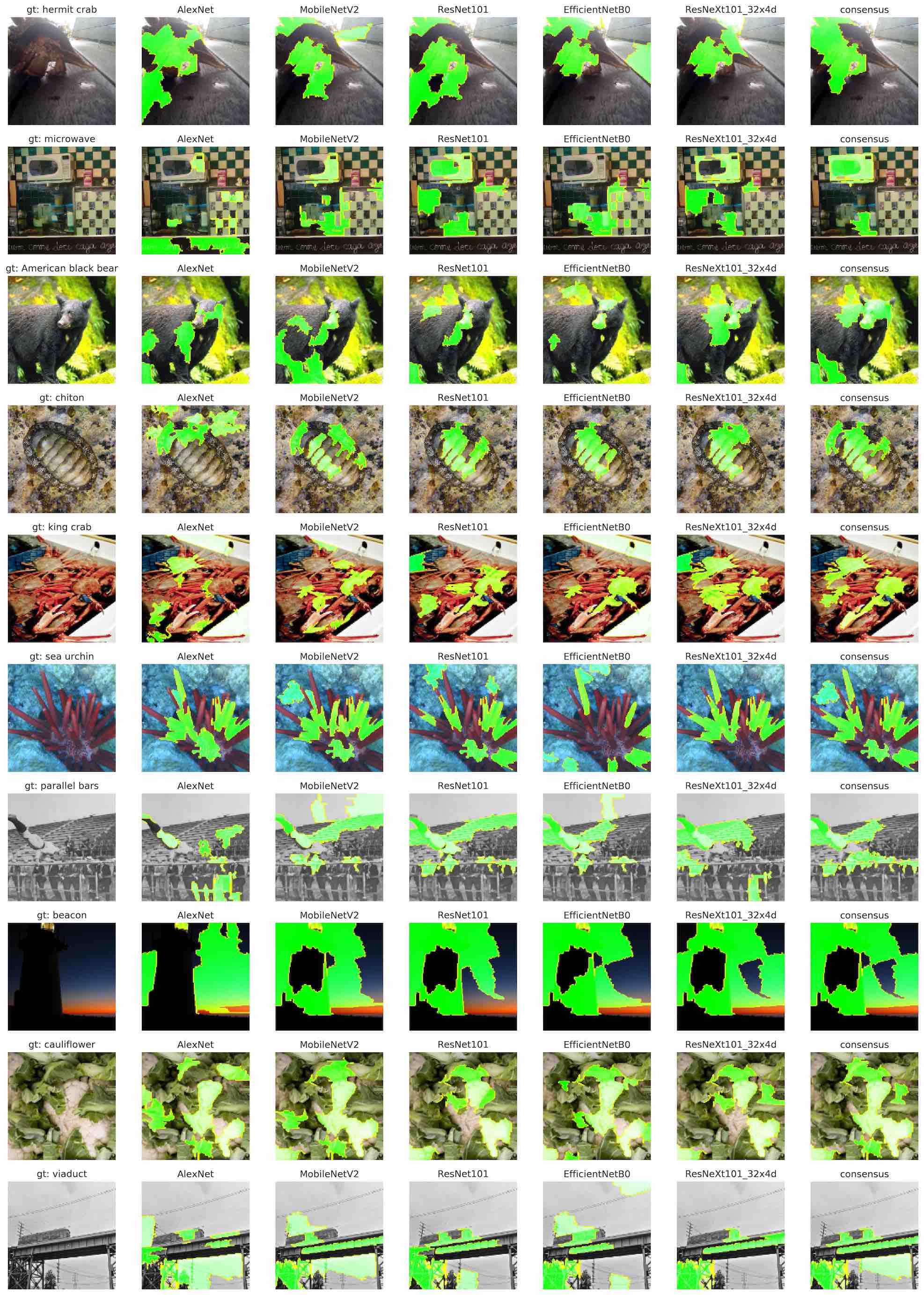}
	    \caption{More visual comparisons between the consensus and the explanations of deep models with LIME on samples from ImageNet. Note that the consensus (last column) is the cross-model consensus of explanations.}
	    \label{fig:imagenet_models_lime_qualitative_more}
	\end{figure*}

\begin{figure*}[t]
	    \centering
	    \includegraphics[width=0.8\linewidth]{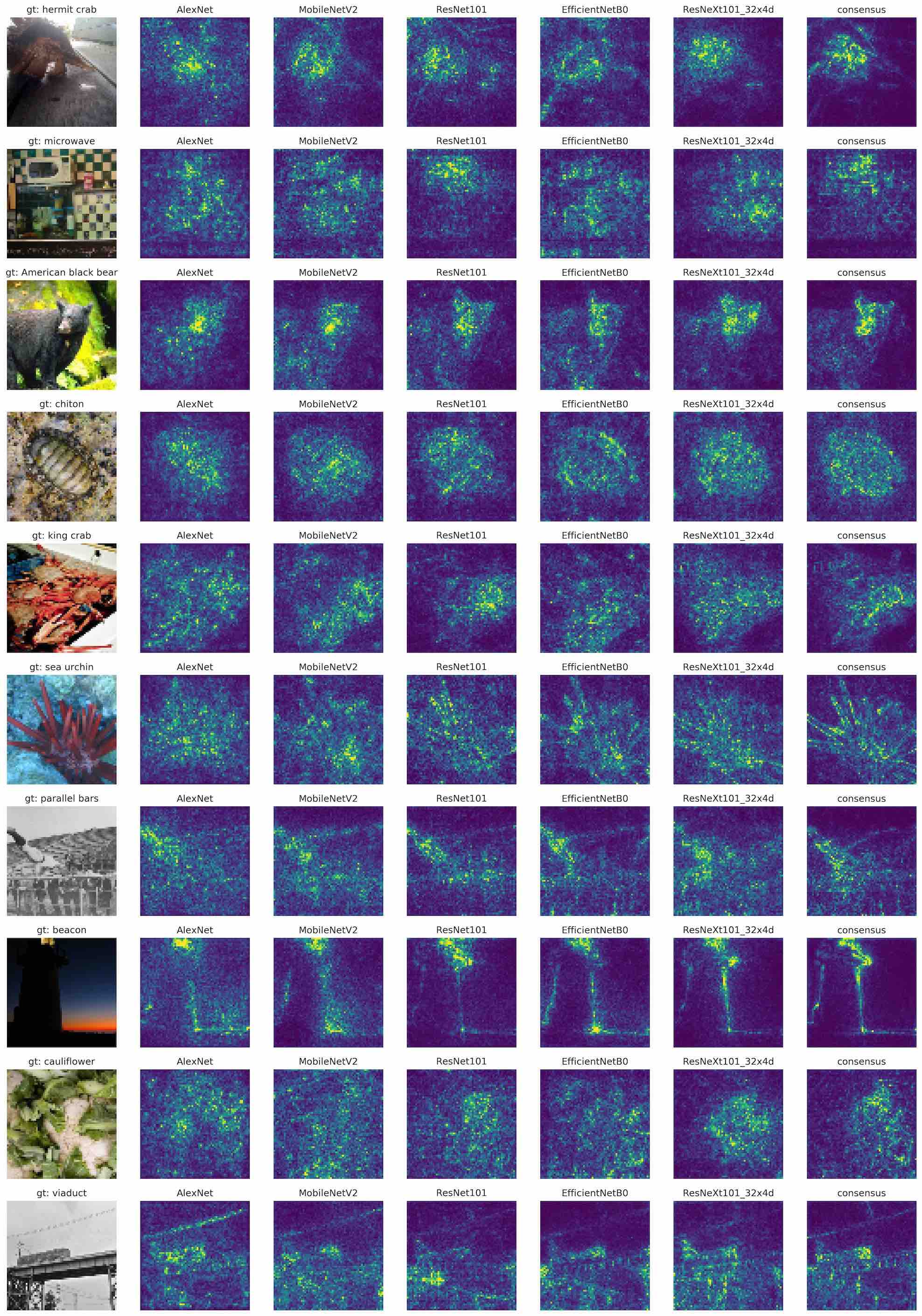}
	    \caption{More visual comparisons between the consensus and the explanations of deep models with SmoothGrad on samples from ImageNet. Note that the consensus (last column) is the cross-model consensus of explanations.}
	    \label{fig:imagenet_models_sg_qualitative_more}
	\end{figure*}

\begin{figure*}[t]
	    \centering
	    \includegraphics[width=0.8\linewidth]{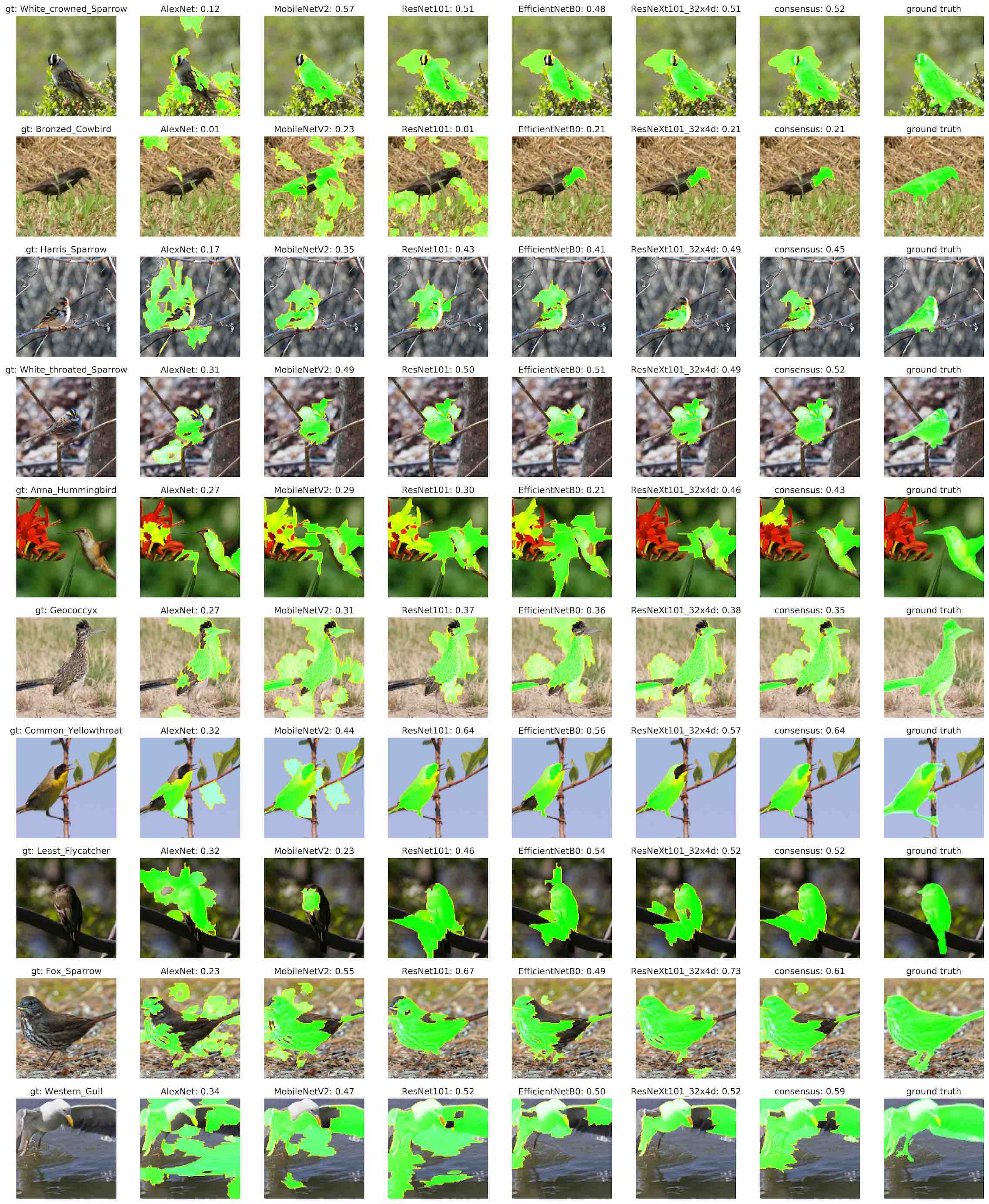}
	    \caption{More visual comparisons between the consensus and the explanations of deep models with LIME on samples from CUB-200-2011, where the pixel-wise annotations of image segmentation are available and the mAPs are measured for the similarity to the segmentation ground truth. Note that the consensus (second last column) is the cross-model consensus of explanations.}
	    \label{fig:cub_models_lime_qualitative_more}
	\end{figure*}

\begin{figure*}[t]
	    \centering
	    \includegraphics[width=0.8\linewidth]{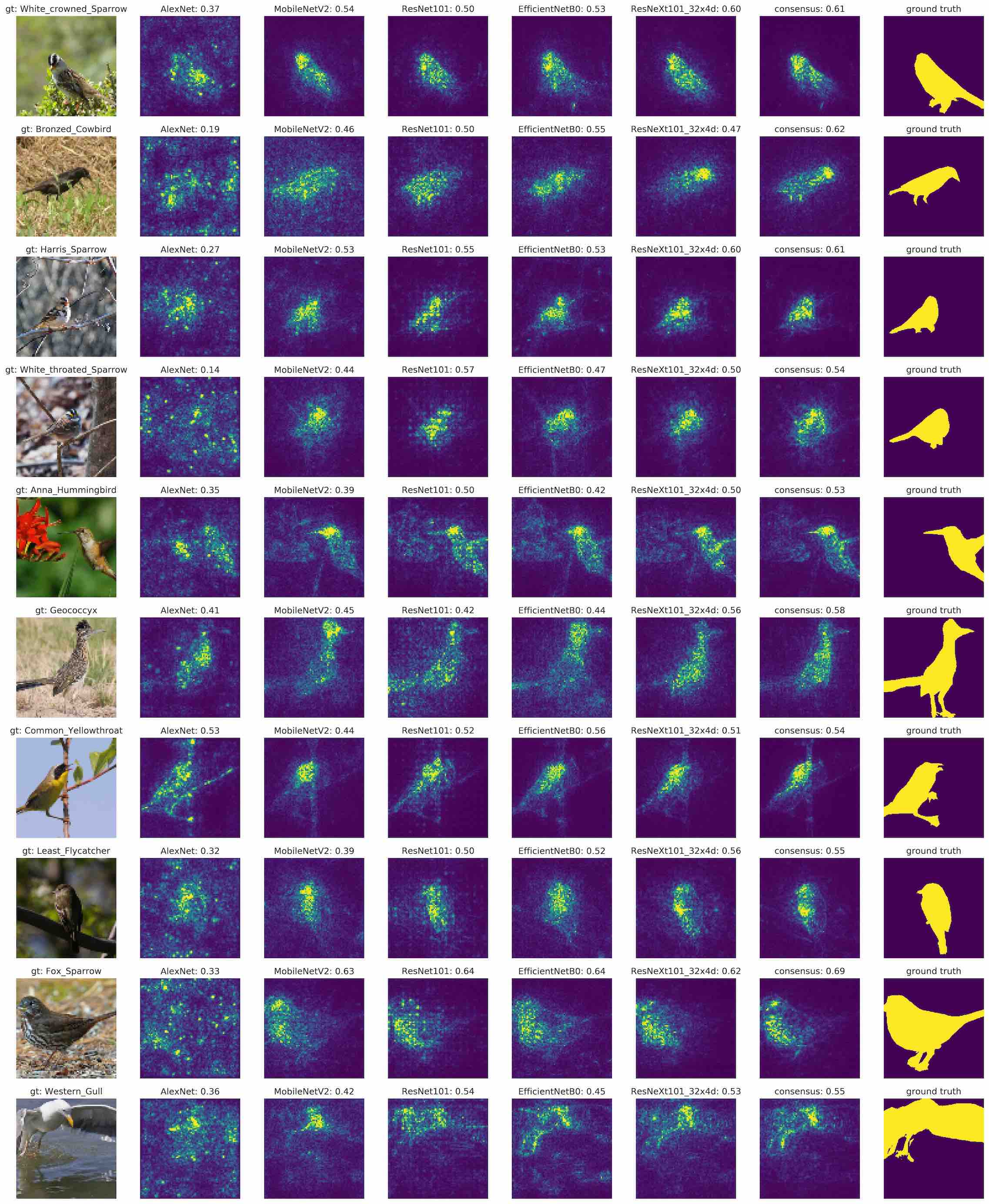}
	    \caption{More visual comparisons between the consensus and the explanations of deep models with SmoothGrad on samples from CUB-200-2011, where the pixel-wise annotations of image segmentation are available and the mAPs are measured for the similarity to the segmentation ground truth. Note that the consensus (second last column) is the cross-model consensus of explanations.}
	    \label{fig:cub_models_sg_qualitative_more}
	\end{figure*}

	We present more visualization results of cross-model consensus of explanations in Figure~\ref{fig:imagenet_models_lime_qualitative_more}, Figure~\ref{fig:imagenet_models_sg_qualitative_more}, Figure~\ref{fig:cub_models_lime_qualitative_more} and Figure~\ref{fig:cub_models_sg_qualitative_more}, where the samples are from ImageNet and CUB-200-2011.
	
\begin{figure*}[t]
		\centering
		\subfloat{\includegraphics[width=0.8\linewidth]{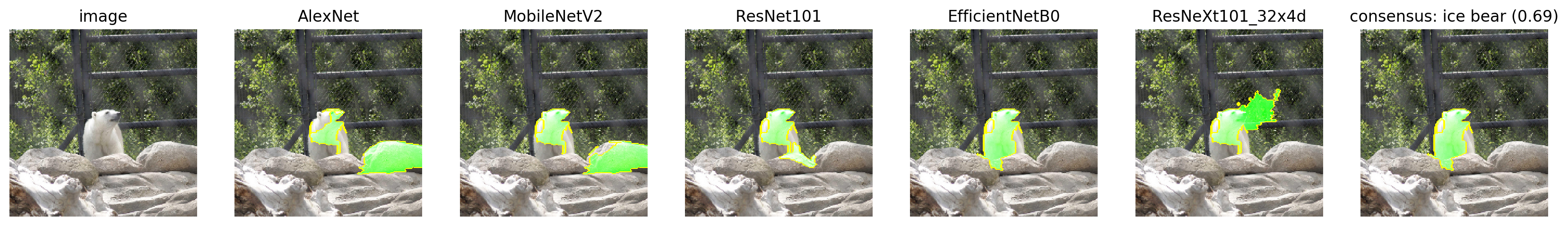}}
		
		\subfloat{\includegraphics[width=0.8\linewidth]{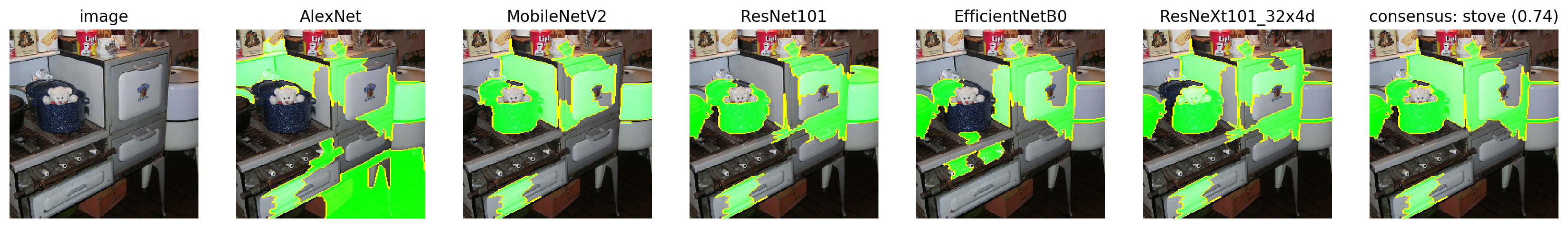}}
		
		\subfloat{\includegraphics[width=0.8\linewidth]{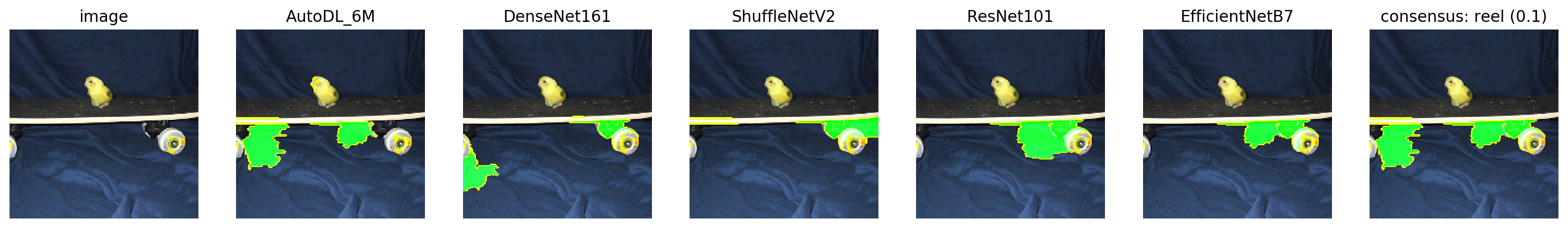}}
		
		\subfloat{\includegraphics[width=0.8\linewidth]{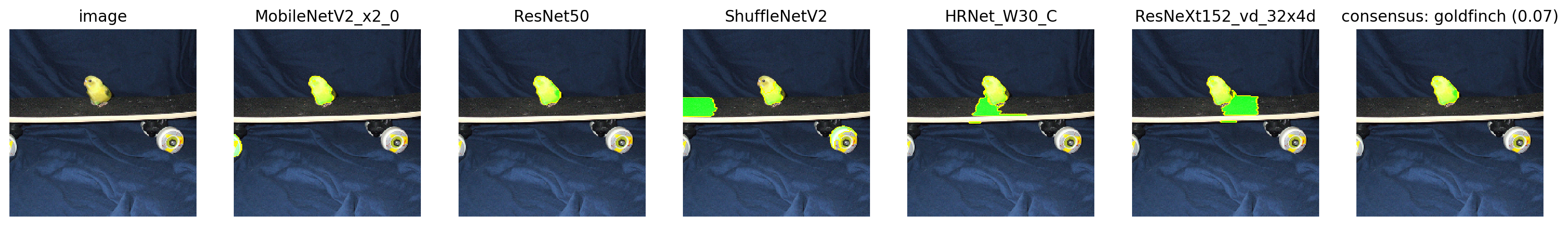}}
		
		\caption{Visualization of images from the MS-COCO dataset \citep{lin2014microsoft} for showing the potentials of cross-model consensus of explanations, where the predicted label with probability is noted.}
		\label{fig:coco}
	\end{figure*}

    For further explorations, we visualize several random images from MS-COCO \citep{lin2014microsoft}, shown in Figure \ref{fig:coco}.
    As introduced in Section~\ref{sec:discussions-consensus}, one direction for the future work would extend the proposed framework on the datasets with multiple visual objects and complicated background for object detection, segmentation, and multi-label classification tasks. 
    
\end{document}